\documentclass[lettersize,journal]{IEEEtran}
\usepackage{amsmath,amsfonts}
\usepackage{algorithmic}
\usepackage{array}
\usepackage[caption=false,font=normalsize,labelfont=sf,textfont=sf]{subfig}
\usepackage{textcomp}
\usepackage{stfloats}
\usepackage{url}
\usepackage{verbatim}
\usepackage{graphicx}
\usepackage{cite}
\usepackage{hyperref}
\usepackage{booktabs}       
\usepackage{nicefrac}       
\usepackage{microtype}      
\usepackage{xcolor}         
\usepackage{multirow}
\usepackage{multicol}
\usepackage{wrapfig}
\usepackage[ruled,linesnumbered,boxed]{algorithm2e}
\usepackage{lipsum}
\usepackage{adjustbox}
\usepackage{subfig}
\usepackage{tabularx}
\usepackage{color}
\usepackage{colortbl}

\begin{document}

\title{Balancing the Causal Effects in Class-Incremental Learning}

\author{Junhao Zheng, Ruiyan Wang, Chongzhi Zhang, Huawen Feng, and Qianli Ma~\IEEEmembership{Member,~IEEE,}
\thanks{Corresponding author: Qianli Ma.}
\thanks{The authors are with the School of Computer Science and Engineering, South China University of Technology, Guangzhou 510006, China (e-mail: junhaozheng47@outlook.com; zswrydyx@163.com; cschongzhizhang@mail.scut.edu.cn; 541119578@qq.com; qianlima@scut.edu.cn).}
\thanks{Manuscript received February 16, 2024.}}

\markboth{Submitted to IEEE Transactions on Neural Networks and Learning Systems}%
{Zheng \MakeLowercase{\textit{et al.}}: A Sample Article Using IEEEtran.cls for IEEE Journals}


\maketitle

\begin{abstract}
Class-Incremental Learning (CIL) is a practical and challenging problem for achieving general artificial intelligence.
Recently, Pre-Trained Models (PTMs) have led to breakthroughs in both visual and natural language processing tasks.
Despite recent studies showing PTMs' potential ability to learn sequentially, a plethora of work indicates the necessity of alleviating the catastrophic forgetting of PTMs.
Through a pilot study and a causal analysis of CIL, we reveal that the crux lies in the imbalanced causal effects between new and old data.
Specifically, the new data encourage models to adapt to new classes while hindering the adaptation of old classes.
Similarly, the old data encourages models to adapt to old classes while hindering the adaptation of new classes.
In other words, the adaptation process between new and old classes conflicts from the causal perspective.
To alleviate this problem, we propose Balancing the Causal Effects (BaCE) in CIL.
Concretely, BaCE proposes two objectives for building causal paths from both new and old data to the prediction of new and classes, respectively.
In this way, the model is encouraged to adapt to all classes with causal effects from both new and old data and thus alleviates the causal imbalance problem.
We conduct extensive experiments on continual image classification, continual text classification, and continual named entity recognition.
Empirical results show that BaCE outperforms a series of CIL methods on different tasks and settings. 
\end{abstract}

 
\section{Introduction}
\IEEEPARstart{I}{ncremental} Learning (IL) aims to endow machine learning systems with the ability to continuously learn novel concepts, which is critical for human-level intelligence research.
This paper focuses on Class-Incremental Learning (CIL), the most challenging and practical scenario in IL \cite{prabhu2020gdumb,buzzega2020dark,van2019three}.
CIL requires models to classify all classes seen so far without task indexes.
While a good number of approaches \cite{kirkpatrick2017overcoming,hu2021distilling,rebuffi2017icarl,hou2019learning,wu2019large} have been proposed in recent years, most of them rely heavily on experience replay \cite{chaudhry2019tiny}, and they suffer from substantial performance deterioration when the replay data is limited or even non-existent.

Recently, Pre-Trained Models (PTMs), especially pretrained Transformers \cite{vaswani2017attention}, have achieved remarkable progress in both computer vision \cite{he2022masked,dosovitskiy2020image} and natural language processing (NLP) \cite{devlin-etal-2019-bert,OpenAI2023GPT4TR}. 
Despite its success across various benchmarks, the CIL ability of pretrained transformers is yet to be fully explored and understood.
On the one hand, the CIL performance of PTMs \cite{wang2022learning,huang2021continual,zheng2022distilling,de2019episodic} is still far from satisfactory with limited buffer data. 
On the other hand, \cite{ramasesh2022effect,taocan} show that PTMs are inherently resilient to catastrophic forgetting even without buffer data.
This contradictory phenomenon urges us to explore the reason behind it.

First, we conduct a pilot study based on linear probing \cite{taocan,chenforgetting} in the CIL scenario.
In the probing study, the PTM is frozen while the classifier is re-trained on the data from all classes learned so far.
Surprisingly, we find that simply retraining the classifier improves the average accuracy \cite{wang2022learning,chaudhry2019tiny} from 14.1\% to 83.2\% without buffer data and from 60.9\% to 84.5\% with 100 old samples in the 20-step setting of split CIFAR-100 \cite{krizhevsky2009learning}.
Then, we track the distance of class centres between PTMs and classifiers.
We find that when the model adapts to new classes, the centres of new classes in the classifier always align with those of PTMs.
In stark contrast, the centres of old classes in classifiers are always pushed away from those of PTMs.
This finding implies that the effects of adapting to new and old classes are contradictory.
Since the new and old data is always imbalanced in CIL \cite{japkowicz2002class,he2009learning} (that is, the class imbalance problem), it exacerbates the causal imbalance problem and leads to catastrophic forgetting even when PTM retain the old knowledge.

Next, we investigate the causal relationship in CIL with causal graphs.
The causal graph is a graphical framework that represents a causal interpretation of data, rather than just a statistical association between them \cite{glymour2016causal,pearl2009causality}.
Specifically, by framing the data, features, and models into causal graphs, we reveal that there are two conflicting causal effects in the adaptation between new and old classes in the data replay setting.
The first causal effect is the effect of new data on the adaptation of old classes.
The second causal effect is the effect of old data on the adaptation of new classes.
In this case, only new or old data has causal effects on adapting to new or old classes, and the model may fail to learn the optimal representation when only limited old data is stored.

To address the problem of causal imbalance, we propose to \textit{\textbf{Ba}lancing the \textbf{C}ausal \textbf{E}ffects} (\textbf{BaCE}) in CIL.
Concretely, BaCE is based on the theory of causal inference and contains two objectives $Effect_{new}$ and $Effect_{old}$.
The former builds balanced causal effects from both new and old data when adapting to new classes.
And the latter builds balanced causal effects from both new and old data when adapting to old classes.
In summary, BaCE encourages models to learn from new and old data in a mutually beneficial way, which mitigates catastrophic forgetting in CIL.

Finally, we conduct extensive experiments on three tasks: continual image classification, continual text classification, and continual named entity recognition.  
The experimental results suggest that BaCE outperforms a series of CIL methods based on ResNet \cite{he2016deep}, e.g.\cite{rebuffi2017icarl,hou2019learning,wu2019large,hu2021distilling}, Vision Transformers \cite{dosovitskiy2020image}, e.g.\cite{wang2022learning}, and BERT \cite{devlin-etal-2019-bert}, e.g.,\cite{de2019episodic,huang2021continual,zheng2022distilling,wang2022less}.
In summary, our contributions are threefold:
\begin{itemize}
    \item We reveal that PTMs are resistant to forgetting, but the causal imbalance problem causes the forgetting of the whole model. 
    \item We delve into the causalities in CIL and discover that the causal effects between new and old data are imbalanced under the experience replay setting. 
    \item We propose BaCE to address the causal imbalance problem and verify its effectiveness on both visual and NLP tasks.
\end{itemize}

\section{Related Work}
\label{sec:related_work}

We summarize four parts of related work: Class-incremental learning, incremental learning with PTM, class-imbalanced problem, causal inference, and continuous causal discovery.

\subsection{Class Incremental Learning} 
Existing CIL methods can be roughly divided into three groups: regularization-based methods, exemplar-based methods, and architecture-based methods. 
\textit{Regularization-based} methods estimate the importance of parameters for previous tasks and penalize the update of important parameters for mitigating forgetting \cite{kirkpatrick2017overcoming,li2017learning,zenke2017continual}.
These methods did not achieve satisfactory performance under challenging and complex scenarios \cite{rebuffi2017icarl}.
\textit{Exemplar-based} methods store representative instances from old classes and replay the stored instances when learning new tasks \cite{rebuffi2017icarl,hou2019learning,wu2019large,buzzega2020dark,arani2022learning}.
Although they achieve state-of-the-art performance on various CIL benchmarks \cite{chaudhry2019tiny}, their performance deteriorates when the buffer size is small.
More importantly, overreliance on exemplars violates the setting of CIL and simplifies CIL as Multi-Task Learning (MTL).
This study investigates practical CIL scenarios where the buffer size is limited or without a rehearsal buffer.
\textit{Architecture-based} methods increase model components incrementally to meet the requirements of new classes \cite{serra2018overcoming,rajasegaran2019random,yan2021dynamically,kim2023stability}.
However, these models require a large memory when there are many tasks.
Furthermore, architecture-based methods implicitly introduce an extra memory budget since the backbones from history are treated as unforgettable checkpoints \cite{zhou2023model}.

Additionally, there are some emerging research directions in CIL.
\cite{wortsman2020supermasks} finds subnetworks for each task during training and infers the task using gradient-based optimization for inference. 
Since the model parameters are not updated in the training, this method does not suffer from catastrophic forgetting.
However, finding subnetworks prohibits the potential forward and backward knowledge transfer in CIL.
\cite{kim2022theoretical} decomposes the CIL problem into within-task prediction and task-id prediction and proposes using out-of-distribution detection techniques to infer the task id.
\cite{kim2022multi} proposes to train a multi-head classifier with an out-of-distribution class for each head and utilizes the adapter to prevent interference between the parameters of different tasks.

\subsection{Incremental Learning with PTMs} 
Most existing Incremental Learning (IL) methods are based on computational neural networks.
Recently, IL with PTMs has become a newly emerging research field.
For example, \cite{wang2022learning,wang2022dualprompt} leverage prompt learning \cite{liu2023pre} for IL and achieve superior performance even without replay data.
However, \cite{wang2022learning,wang2022dualprompt} introduce an extra architecture called prompt pool.
Unlike \cite{wang2022learning,wang2022dualprompt}, BaCE does not rely on additional components.
In addition, \cite{ermis2022memory,razdaibiedinaprogressive} utilize adapter \cite{houlsby2019parameter} and prompt, respectively, to solve Task-Incremental Learning, which is an easier scenario than CIL since task indexes are provided during inference.
\cite{wangs2022sprompt} learns prompts independently across domains for Domain-Incremental Learning. 
Moreover, \cite{ke2022adapting,ke2022continual,jang2022towards} focus on continual pretrainingpretraining with PTMs, which is a more general scenario of continual learning.

There are some other task-specific CIL methods in natural language processing using PTMs, such as BERT.
\cite{wang2020efficient} improves MBPA++ \cite{de2019episodic} and proposes a meta-lifelong framework for text classification and question answering.
\cite{sunlamol} utilizes the pretrained knowledge in gpt2. \cite{radford2019language} and generates pseudo-samples of old tasks when learning new tasks.
\cite{xia2022learn} proposes a two-stage framework Learn-and-Review for continual NER.
It utilizes knowledge distillation and generates pseudo-samples of old entity types when learning new tasks.
\cite{zhang-etal-2023-continual} improves the knowledge distillation to take advantage of existing translation models.
\cite{xia-etal-2023-enhancing} proposes to split the last layer into previous and current classifiers to mitigate the bias of the classifier and the representation for continual relation extraction. 

\subsection{Class imbalance problem}
The class imbalance problem \cite{japkowicz2002class,he2009learning} between old and new classes is a long-standing problem in CIL, and many studies have attempted to alleviate it.
For example, LUCIR \cite{hou2019learning} proposes using a cosine classifier to avoid the imbalanced magnitudes between the new and old predictions.
IL2M \cite{belouadah2019il2m} introduces an additional memory to store the statistics of old tasks obtained when they were first learned.
BiC \cite{wu2019large} addresses the data imbalance between the old and new classes by finetuning classifiers on balanced data.
The causal imbalance problem is closely related to the class imbalance problem.
The key difference is that the causal imbalance problem describes the learning process at the feature level while the class imbalance problem describes the final prediction at inference time.

\subsection{Causal Inference in CV and NLP}
Causal inference \cite{glymour2016causal,pearl2009causality} has recently been introduced to various visual and NLP tasks such as image classification \cite{hu2021distilling}, long-tailed classification \cite{tang2020long,nan2021uncovering}, distantly supervised named entity recognition \cite{zhang2021biasing}, neural dialogue generation \cite{zhu2020counterfactual}, continual named entity recognition \cite{zheng2022distilling}, and finetuning \cite{zheng2023preserving}.
Our idea stems from the causal view of forgetting in \cite{hu2021distilling}, and the proposed BaCE further seeks a balance between the causal effects of new and old data. 

\subsection{Continual Causal Discovery}
Causality theory \cite{pearl2009causality} provides language, algorithms, and tools to discover and infer cause-and-effect relationships from any collection of observational/experimental data based on a partial understanding of a complex system.
Despite huge strides in causality in recently years, few studies consider the continual learning setting in causal discovery \cite{mundt2023continual}.
\cite{javed2020learning,gong2023active} focused on learning causal structures from a data stream.
Unlike them, our research focuses on class-incremental learning instead of finding causal structures behind data.

\section{A Pilot Study for CIL with PTMs}
In this section, we analyze the key to PTMs' forgetting.
Typically, a model can be divided into two components: a feature encoder and a task-specific classifier.
Without loss of generality, we use PTMs as the feature encoder and a linear layer with cosine normalization \cite{hou2019learning} as the classifier.

\begin{figure}[!t]
    \centering
    \subfloat[SEQ]{
        \includegraphics[width=0.30\linewidth]{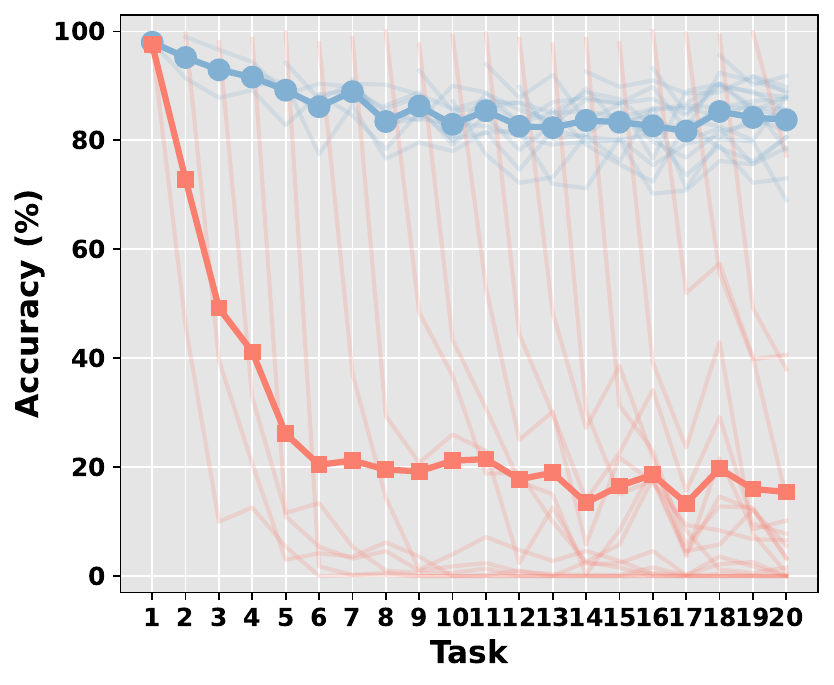}
        \label{fig:probing_vit_a}
    }
     \subfloat[REPLAY]{
        \includegraphics[width=0.30\linewidth]{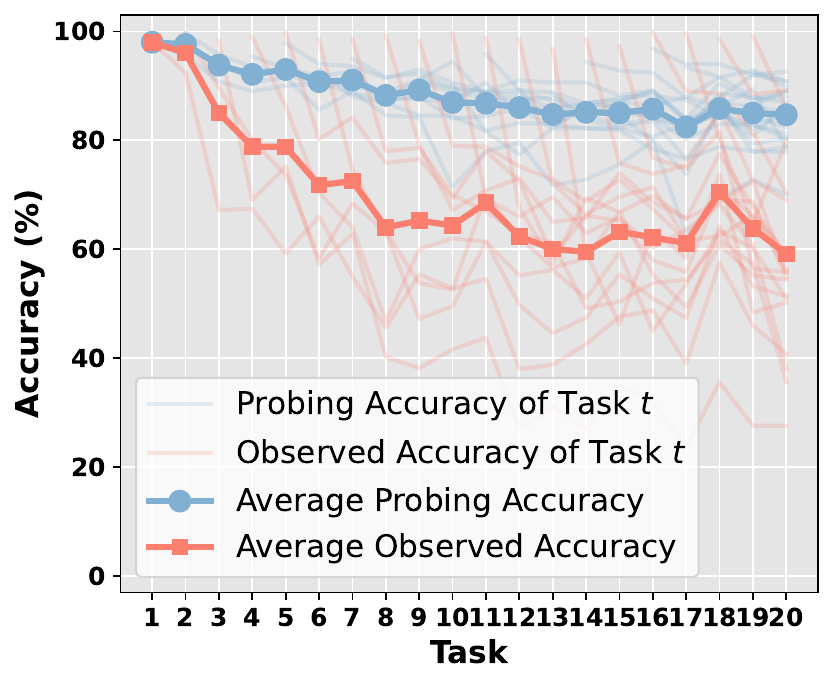}
        \label{fig:probing_vit_b}
    }
     \subfloat[MTL]{
        \includegraphics[width=0.30\linewidth]{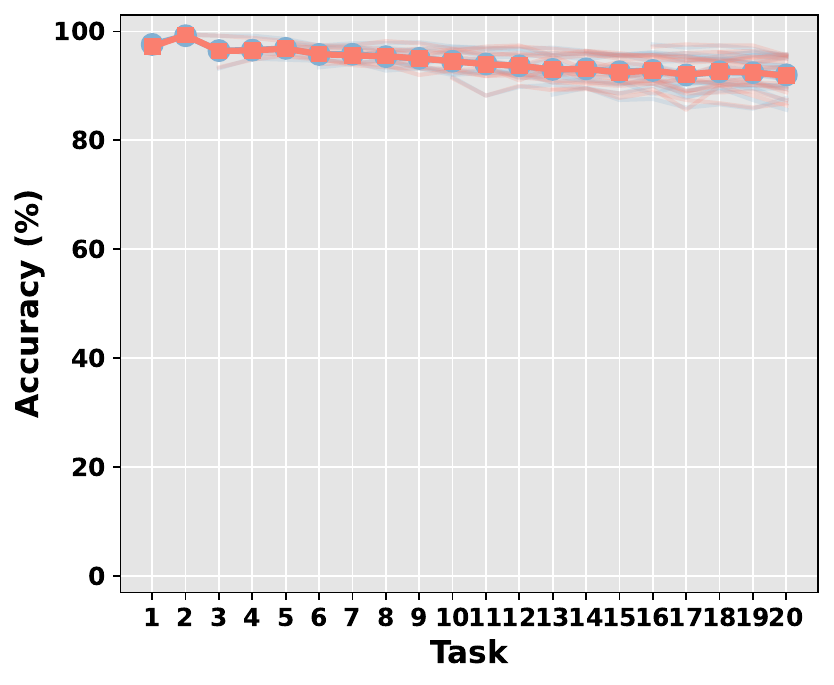}
        \label{fig:probing_vit_c}
    }
    \caption{The probing study on the 20-step split CIFAR-100. We use ViT-B/16 (ViT) pretrained on ImageNet-21k \cite{dosovitskiy2020image} as backbones. The buffer size is 200 in REPLAY. The blue curve represents the observed accuracy and the red curve represents the probing accuracy.}
    \label{fig:probing_vit}
\end{figure}

\subsection{Probing Study}
The linear probing \cite{taocan,chenforgetting} is a commonly used technique to measure the encoding ability of feature encoders.
In the probing study, we aim to probe each model checkpoint in CIL.
Specifically, we fix the encoder of each checkpoint and retrain its classifier on the data of all tasks that have been learned so far.
In this way, we obtain the probing performance of each checkpoint, and this performance can be regarded as the upper limit performance when classifiers do not forget.
Correspondingly, we call the performance of the original model as \textit{the observed performance}.
We note that the retrained new classifier is only for estimation of the probing performance and will not participate in the subsequent training in continual learning.

We consider three methods for probing: sequential training (SEQ), experience replay (REPLAY), and multi-task learning (MTL).
The result of the probing study on split CIFAR100 \cite{krizhevsky2009learning} is shown in Figure \ref{fig:probing_vit}.
Figure \ref{fig:probing_vit} shows that pretrained encoders are resistant to forgetting while trained-from-scratch classifiers are prone to forgetting.
Besides, more rehearsal data helps close the gap between observed and probing performance. 

\begin{figure}[!t]
    \centering
    \subfloat[SEQ]{
        \includegraphics[width=0.30\linewidth]{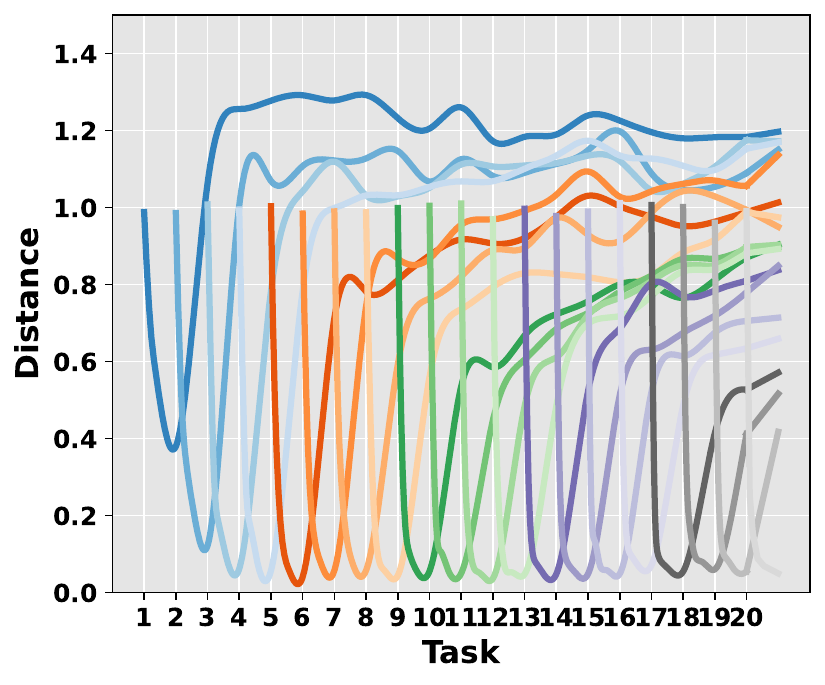}
        \label{fig:tracking_vit_a}
    }
     \subfloat[REPLAY]{
        \includegraphics[width=0.30\linewidth]{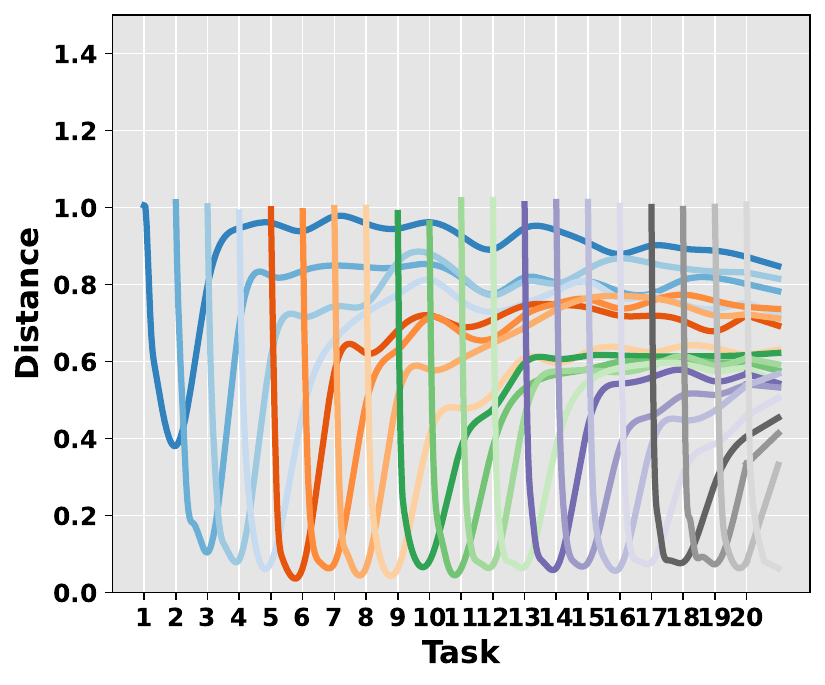}
        \label{fig:tracking_vit_b}
    }
    \subfloat[MTL]{
        \includegraphics[width=0.30\linewidth]{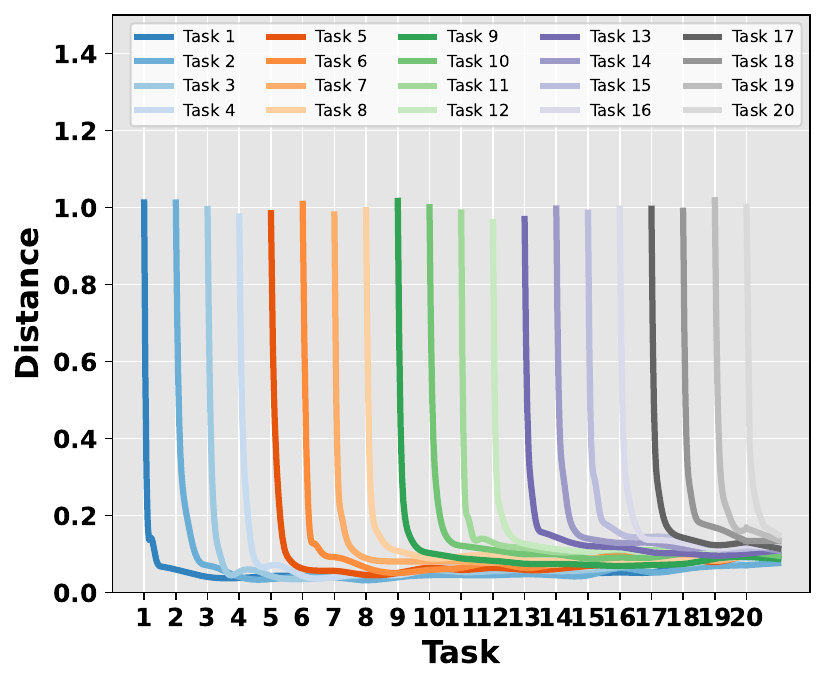}
        \label{fig:tracking_vit_c}
    }
    \caption{The evolution of \textit{feature-embedding distance}. The backbone model is ViT-B/16, and the dataset is the 20-step split CIFAR-100. Each colour represents the average \textit{feature-embedding distance} of classes from an incremental task.}
    \label{fig:tracking_vit}
\end{figure}

\begin{figure}[!t]
    \centering
    \subfloat[Average Accuracy]{
    \includegraphics[width=0.45\linewidth]{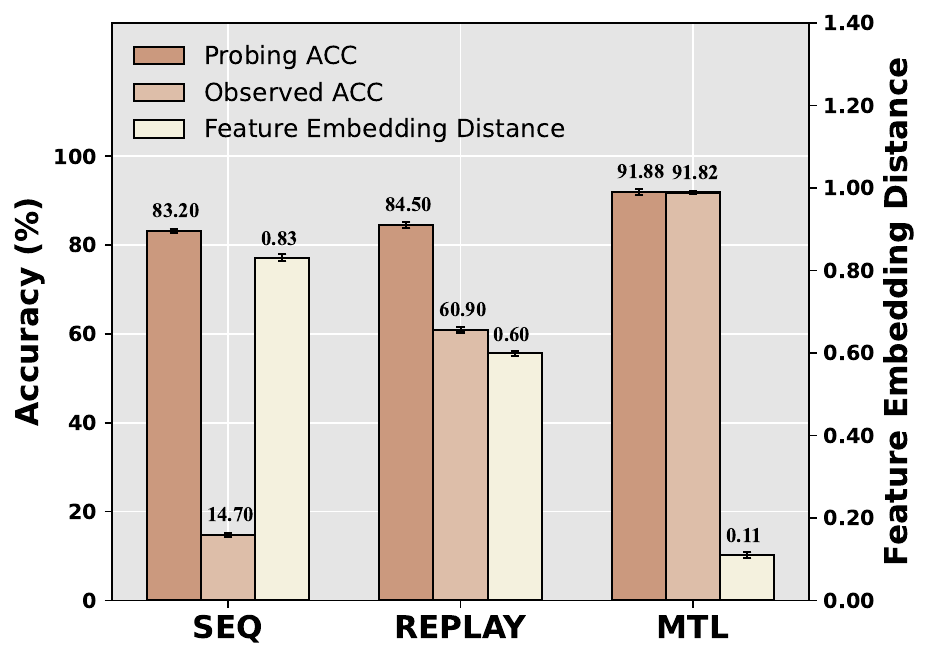}
    \label{fig:tracking_performance_gap_vit_a}
    }
    \subfloat[Forgetting]{
    \includegraphics[width=0.45\linewidth]{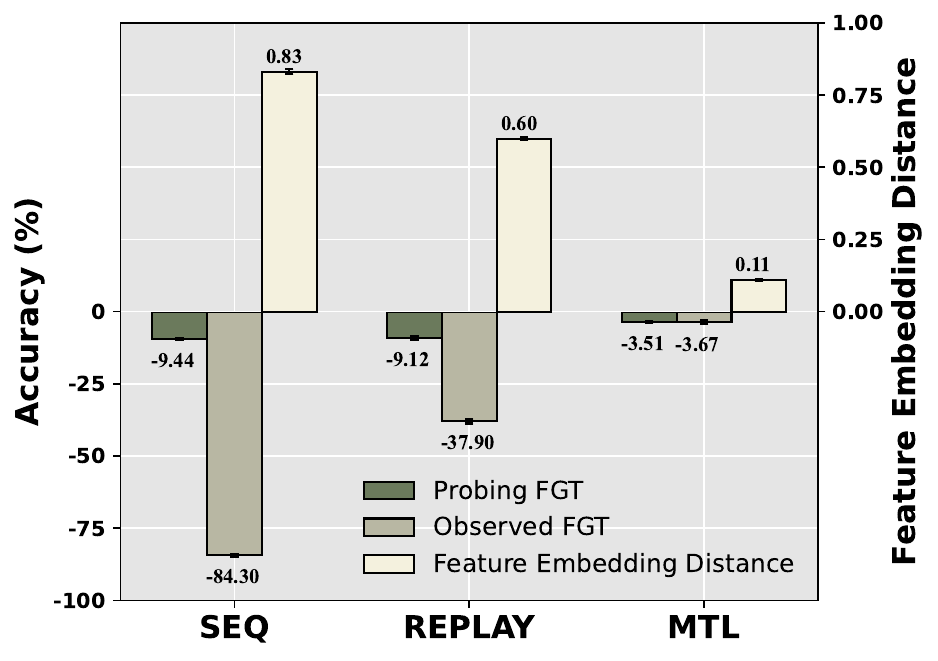}
    \label{fig:tracking_performance_gap_vit_b}
    }
    
    \subfloat[Old and New Tasks]{
    \includegraphics[width=0.90\linewidth]{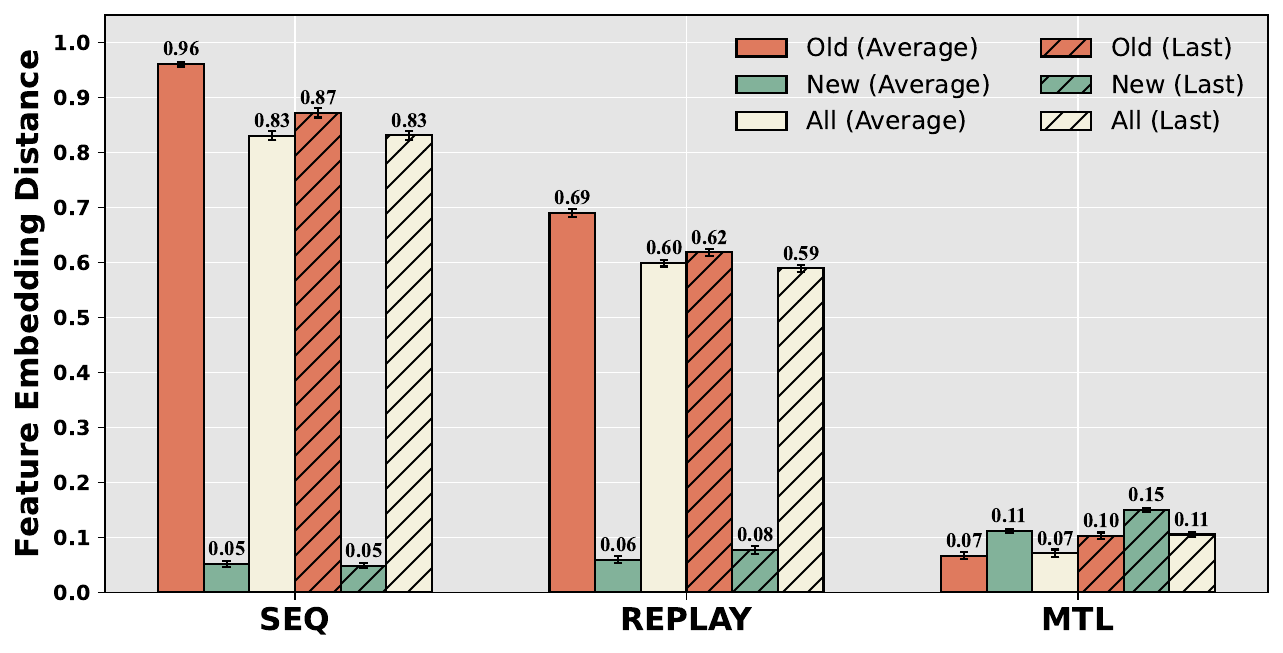}
    \label{fig:tracking_performance_gap_vit_c}
    }
    \caption{(a) and (b) show the relationship between feature-embedding distance and average accuracy and forgetting. (c) shows the feature-embedding distance of old and new tasks. ``Average'' represents the distance averaged over all incremental steps; ``Last'' represents the distance measured at the last incremental step; ``New'', ``Old'', and ``All'' represent the distance of new, old, and all classes, respectively.}
    \label{fig:tracking_performance_gap_vit}
\end{figure}

\begin{figure}[!t]
    \centering
    \includegraphics[width=0.9\linewidth]{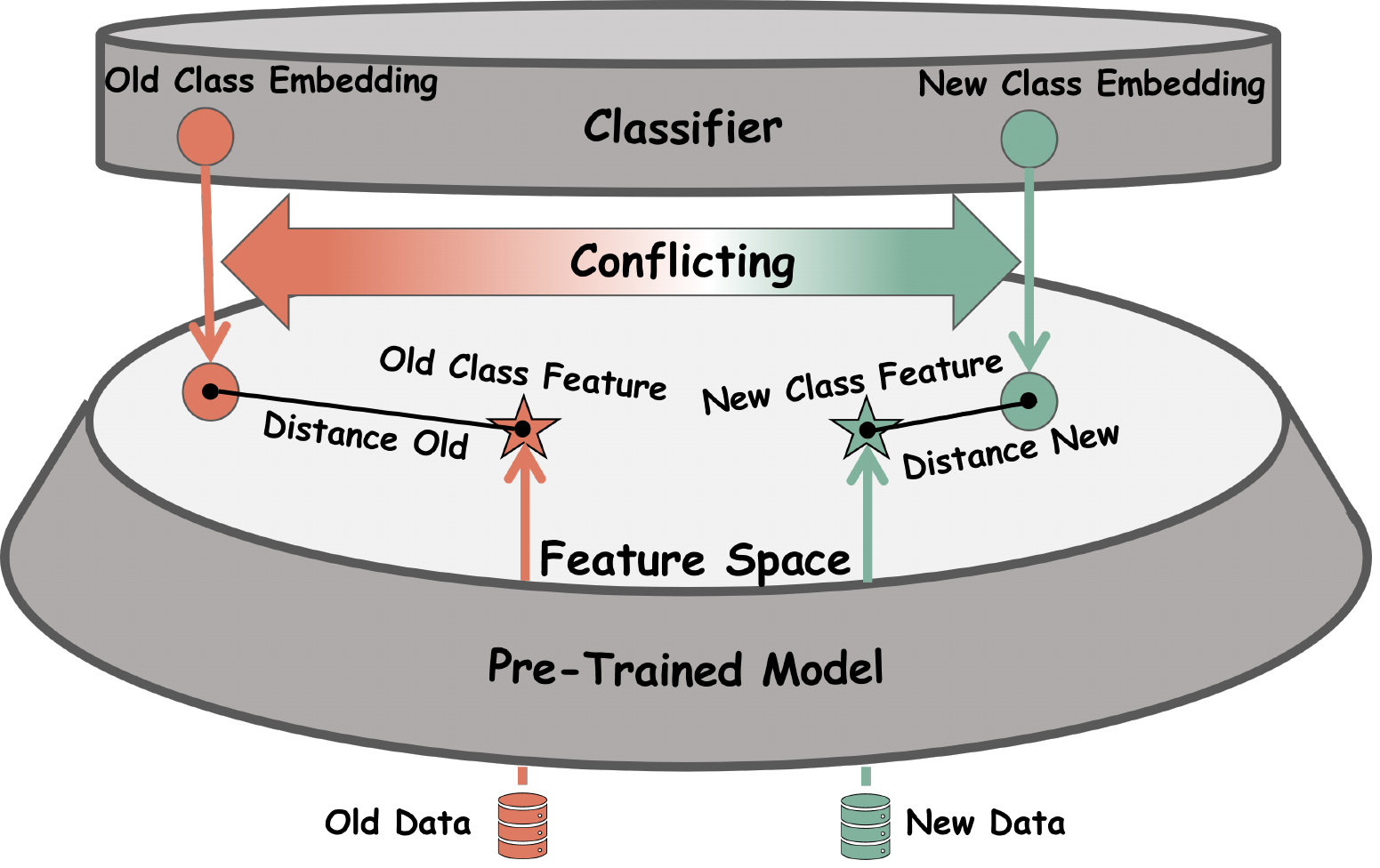}
    \label{fig:illustration_tracking}
    \caption{The illustration of the conflicting causal effects in CIL.}
\end{figure}

\subsection{Tracking Study}
The probing study shows that classifiers and encoders always forget at a different speed.
To understand why it happens, we track the learning process of classifiers and encoders from a feature-level perspective.
In classifiers, the representation of each class (\textit{class embedding}) corresponds to a row vector in the weight matrix.
In encoders, the representation of each class (\textit{class feature}) can be estimated as the average feature of all training samples from that class.
We call the distance between class feature and class embedding the feature-embedding distance.
Intuitively, the feature-embedding distance of one class is small when models learn how to distinguish it from others, and the feature-embedding distance of one class is large when models fail to do so.
Therefore, we can adopt the feature-embedding distance as an indicator of how the conflicting causal effects hinder the learning process in CIL.

Figure \ref{fig:tracking_vit} shows that the feature-embedding distance of each task decreases to a small value when this task is newly adapted, and the distance increases as models learn more tasks.
For each CIL step, new classes are learned as their feature-embedding distances are minimized. 
In contrast, old classes are forgotten as their feature-embedding distances grow.
In other words, models always align new class embeddings with the corresponding new class features while simultaneously pushing old class embeddings away from the corresponding old class features.
Similarly, new classes will be forgotten if new models are trained only on old data. 
In summary, the new or old data hinders the adaptation of old or new classes, suggesting there are conflicting causal effects in the adaptation of new and old classes. 
We illustrate the conflicting causal effects in \ref{fig:illustration_tracking}. 
Alleviating the conflicting causal effects is important because it may hinder models from learning the optimal representations for new and old data.

Figure \ref{fig:tracking_performance_gap_vit_a} shows that when more replay data is available, the feature-embedding distance decreases, and the gap between probing and observed performance is narrowed.
It indicates that minimizing the feature-embedding distance may close the performance gap.
Figure \ref{fig:tracking_performance_gap_vit_b} shows a similar trend from the perspective of forgetting.
Figure \ref{fig:tracking_performance_gap_vit_c} shows that, when training with more replay data, the feature-embedding distance of new tasks increases while that of old tasks dramatically decreases.
It indicates that the new and the old tasks are trained in a confrontational manner caused by the conflicting causal effects.
The conflicting causal effects will not hurt the performance when all new and old data are trained jointly,i.e., the MTL setting.
However, when only limited old data are stored, the conflicting causal effects hinder models from learning the optimal representations of both new and old data.
How can we eliminate the conflicting causal effects when storing limited old samples? 
To answer this question, we need to first sort out the causal relationships in CIL.

\section{Methodology}

We revisit the causal relationship in CIL in Section \ref{sec:revisiting_the_causailities_in_CIL} and reveal that the causal effects of adaptation of new and old classes are imbalanced.
Then, we provide an overview of the proposed BaCE in Section \ref{sec:bace_all} and a detailed description of learning old and new classes with balanced causal effects in Section \ref{sec:bace_effect_old} and \ref{sec:bace_effect_new}, respectively.

\subsection{Revisiting the Causalities in CIL}
\label{sec:revisiting_the_causailities_in_CIL}

Formally, the goal of CIL is to learn a single model $f_\theta:\mathbf{x}\rightarrow y \in \mathcal{Y}$ from the sequence of tasks $\mathcal{D}=\{\mathcal{D}_1,\mathcal{D}_2,\cdots,\mathcal{D}_T\}$, where the $t$-th task $\mathcal{D}_t = \{ (\mathbf{x}_i^t,y_i^t)\}_{i=1}$ contains input samples $\mathbf{x}_i^t \in \mathcal{X}_t$ and labels $y_i^t \in \mathcal{Y}_t$.
The label sets of different tasks are exclusive: $\mathcal{Y}_1 \cap \mathcal{Y}_2 \cdots \mathcal{Y}_T = \emptyset$.
When adapting to each new task, the classifier expands the output dimension for predicting new categories.
In the data replay setting, a buffer $\mathcal{M}$ is introduced for storing old representative instances.

Each CIL step can be framed into a causal graph \cite{pearl2009causality}, where nodes are variables and directed edges represent the causalities between variables. 
Figure \ref{fig:method_seq} is the causal graph of SEQ. 
In Figure \ref{fig:method_seq}, $X^{old}$,$X^{new}$ are the input samples from old and new tasks, and $H^{old}$, $H^{new}$ are the extracted features, respectively.
$Z$ represents output logits, i.e., the model predictions before the softmax layer.
The superscript ``old'' and ``new'' of $Z$ represents they are computed from $X^{old}$ and $X^{new}$ respectively.
Moreover, the subscripts ``[old]'' and ``[new]'' represent the logits over the category from old and new tasks.

Typically, the classification loss is computed as the cross-entropy loss of logits.
Therefore, optimizing the logits of new ($Z_{[new]}$) and old classes ($Z_{[old]}$) encourages the adaptation of new and old classes, respectively.
In Figure \ref{fig:method_seq},  although $X^{new}$ have effects on both $Z_{[new]}^{new}$ and $Z_{[old]}^{new}$ in forward propagation, only the causal path $X^{new} \rightarrow H^{new} \rightarrow Z_{[new]}^{new}$ helps models adapt to new classes (the {\color{teal} green} path in Figure \ref{fig:method_seq}) while the other path $X^{new} \rightarrow H^{new} \rightarrow Z_{[old]}^{new}$  (the {\color{black}\textbf{black}} path in Figure \ref{fig:method_seq}) hinders this process when considering backward propagation.
For clarity, we call the causal effects that enhance the adaptation of new classes as the \textbf{positive causal effects} and the causal effects that hinder the adaptation of old classes as the \textbf{conflicting causal effects}.
When models only adapt to new classes, all training samples are from new classes, and thus, only \textbf{positive causal effects} exists.

Next, let us consider the data replay setting.
The causal graph of REPLAY is shown in Figure \ref{fig:method_replay}, where $X^{buf}$ is the rehearsal samples selected from $X^{old}$.
Apart from the causal path of adapting new classes (the {\color{teal} green} path in Figure \ref{fig:method_replay}), there is another causal path $X^{old} \rightarrow X^{buf} \rightarrow H^{buf} \rightarrow Z_{[old]}^{buf}$ (the {\color{red!50} red} path in Figure \ref{fig:method_replay}) adapting models to old classes.
For Figure \ref{fig:method_replay}, both {\color{teal} green} and {\color{red!50} red} paths are positive causal effects since they enhance adaptation to new and old classes respectively.

In addition to the two causal paths, we find that the {\color{black}\textbf{black}} paths in Figure \ref{fig:method_replay} hinder the adaptation of the other classes.
Specifically, the causal path $H^{buf} \rightarrow Z_{[new]}^{buf}$ hinders the adaptation of new classes and $H^{new} \rightarrow Z_{[old]}^{new}$ hinders the adaptation of old classes.
We call these two causal effects the \textbf{conflicting causal effects}.
Suppose that if we store all old data in REPLAY, the conflicting causal effects will cancel out.
However, in the practical CIL scenario, only a small number of old data can be stored as buffer data.
In this case,  the conflicting causal effects may not cancel out since the buffer data may not represent the true distribution of old data.
A straightforward solution is to customize the weight of cross-entropy loss of different classes.
However, it requires a lot of manual efforts to search the hyperparameters and is unapplicable in the real-world scenario.

\subsection{BaCE: Balancing the Causal Effects in CIL}
\label{sec:bace_all}

\begin{figure*}[!t]
    \centering
    \subfloat[SEQ]{
        \includegraphics[width=0.12\linewidth]{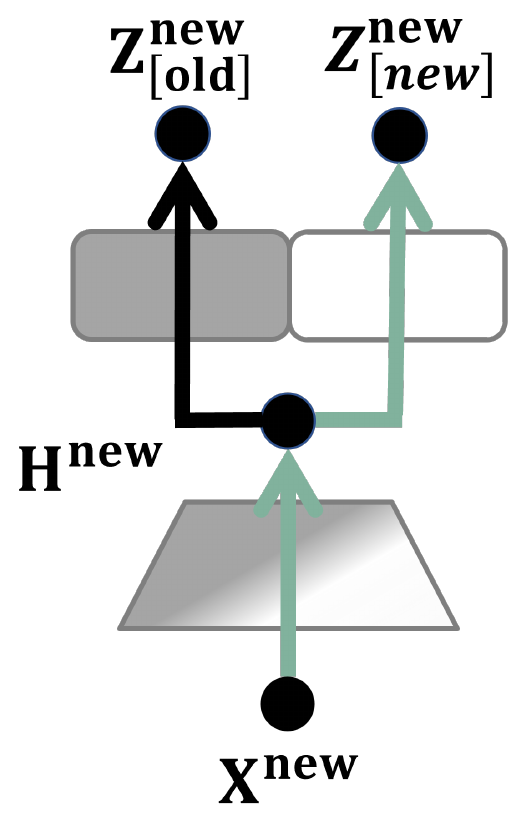}
        \label{fig:method_seq}
    }
     \subfloat[REPLAY]{
        \includegraphics[width=0.17\linewidth]{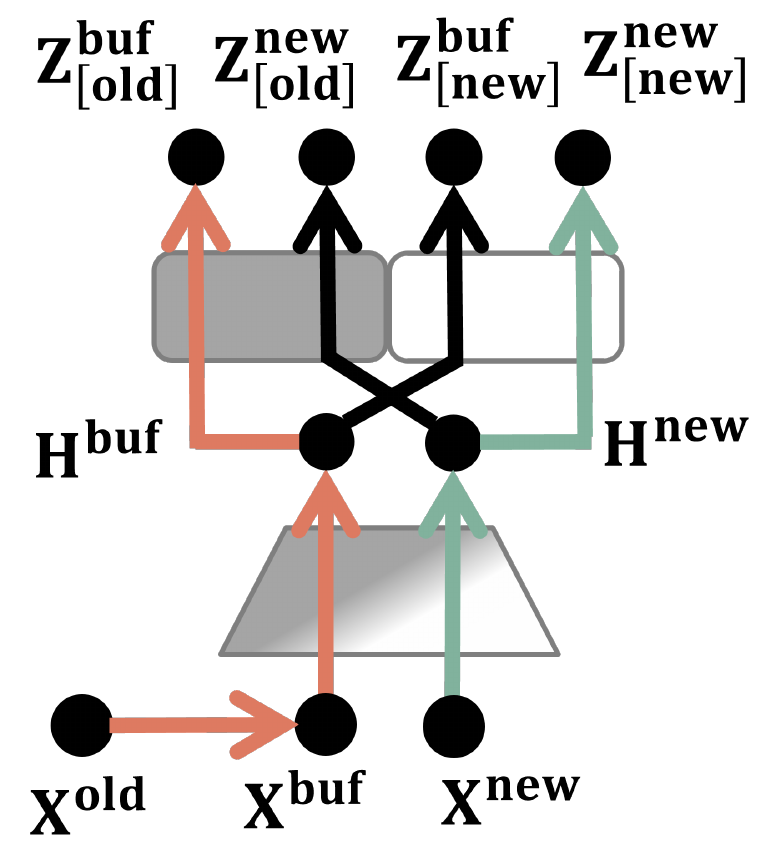}
        \label{fig:method_replay}
    }
    \subfloat[BaCE w/ $\textit{Effect}_{old}$]{
        \includegraphics[width=0.20\linewidth]{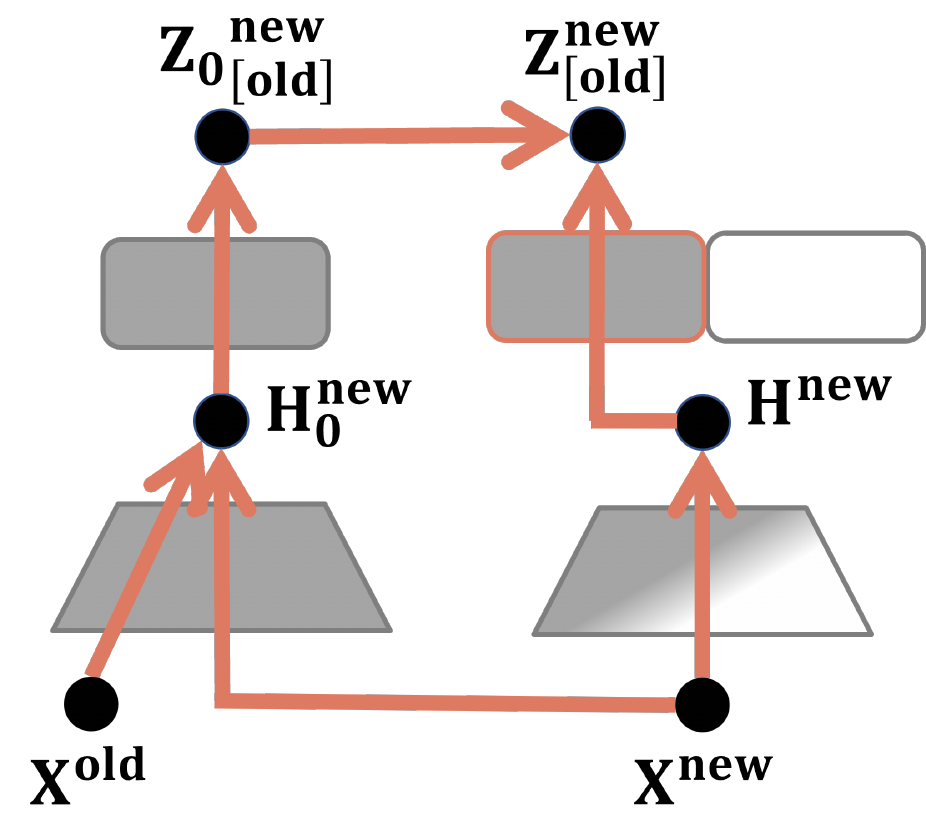}
        \label{fig:bace_effect_old}
    }
    \subfloat[BaCE w/ $\textit{Effect}_{new}$]{
        \includegraphics[width=0.20\linewidth]{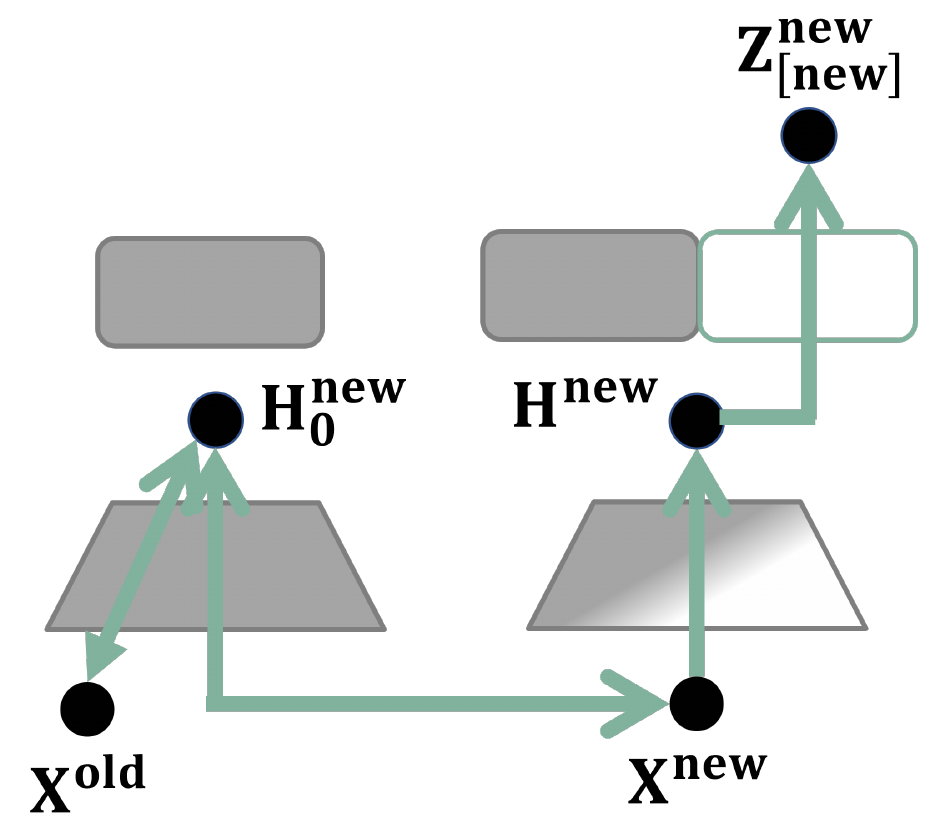}
        \label{fig:bace_effect_new}
    }
    \subfloat[BaCE]{
        \includegraphics[width=0.21\linewidth]{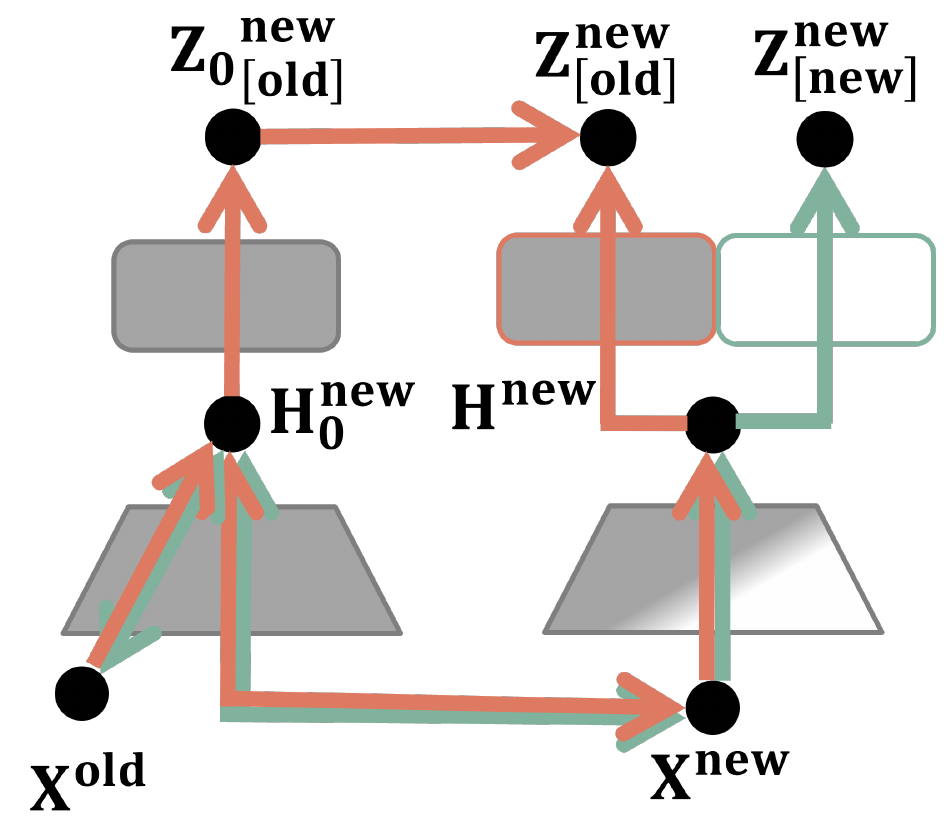}
        \label{fig:bace_effect_new_old}
    }
    \\
    \subfloat[Comparison Between SEQ, REPLAY and BaCE]{
        \includegraphics[width=0.60\linewidth]{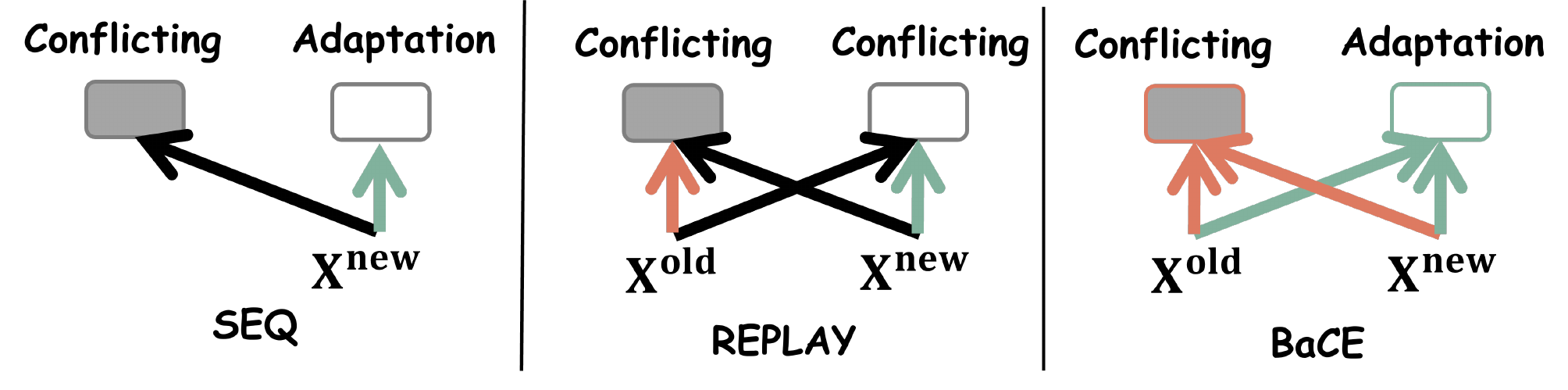}
        \label{fig:method_diff}
    }
    \includegraphics[width=0.19\linewidth]{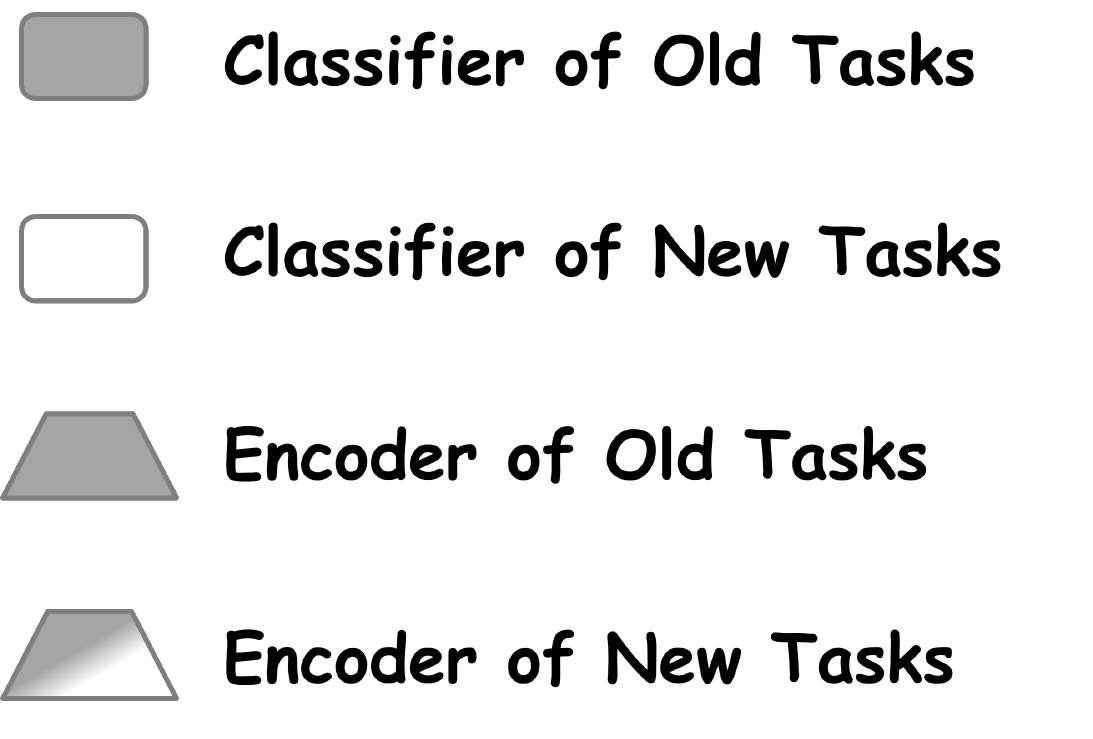}
    \includegraphics[width=0.19\linewidth]{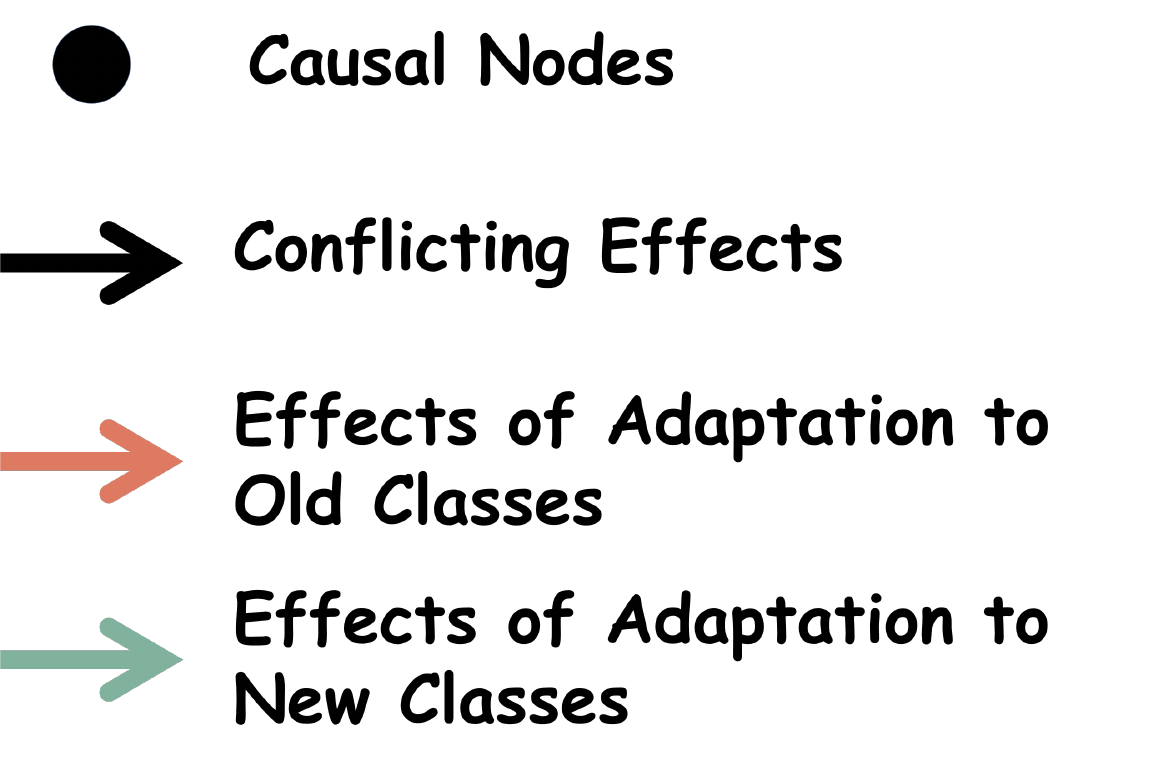}
\caption{The causal graphs of SEQ, REPLAY, and BaCE (Ours) with $\textit{Effect}_{old}$ and $\textit{Effect}_{new}$ in each CIL step. 
    The directed edges represent the causal effects between variables. 
    The {\color{red!50} red} and {\color{teal} green} paths represent the positive causal effects of adaptation of \textit{old} and \textit{new} classes, respectively. 
    The {\color{black} \textbf{black}} paths represent the conflicting causal effects.
    } 
    \label{fig:method}
\end{figure*}

Section \ref{sec:revisiting_the_causailities_in_CIL} reveals the conflicting causal effects in SEQ and REPLAY.
To fundamentally overcome this problem, we propose BaCE, which encourages adaptation to new and old data in a collaborative manner.
Our key intuition is to build positive causal paths from both $X^{old}$ and $X^{new}$ to the logits of old or new classes. 
In other words, we expect models to learn to adapt to either old or new classes with positive causal effects from both $X^{old}$ and $X^{new}$.
The difference between SEQ, REPLAY, and BaCE is illustrated in Figure \ref{fig:method_diff}.

To begin with, we provide an overview of BaCE in Figure \ref{fig:bace_effect_new_old}.
There are two models in BaCE: the teacher model trained on previous tasks (denoted as $f^{t-1}$) and the student model for adapting to the new task (denoted as $f^{t}$).
Different from most existing CIL methods \cite{wu2019large,hou2019learning,buzzega2020dark,li2017learning} based on knowledge distillation \cite{hinton2015distilling}, the teacher model of BaCE is updated every epoch to facilitate the adaptation to new data distribution:
\begin{equation}
    f^{t-1}=\beta f^{t-1}+(1-\beta)f^{t}
\end{equation}
We empirically find that $\beta=0.9$ yields better performance.

Then, we denote the overall objective as \textit{Effect}.
To learn each class with positive causal effects from both $X^{old}$ and $X^{new}$, we decompose the overall objective as follows:
\begin{equation}
    \textit{Effect} = \textit{Effect}_{new} + \textit{Effect}_{old},
\end{equation}
where $\textit{Effect}_{new}$ represents the positive causal effects of adapting to new classes and  $\textit{Effect}_{old}$ represents the positive causal effects of adapting to old classes.
We will introduce how to estimate $\textit{Effect}_{old}$ and $\textit{Effect}_{new}$ in Section \ref{sec:bace_effect_old} and \ref{sec:bace_effect_new}, respectively.

\subsection{$\textit{\textbf{Effect}}_{\textbf{old}}$: Learning Old Classes with Balanced Causal Effects from $X^{old}$ and $X^{new}$.}
\label{sec:bace_effect_old}

We propose to learn old classes with positive and balanced causal effects from $X^{old}$ and $X^{new}$.
The causal graph of $\textit{Effect}_{old}$ is shown in Figure \ref{fig:bace_effect_old}.
Its rationale is as follows: 
$H_0^{new}$ is the feature of $X^{new}$ extracted by the teacher model.
$H_0^{new}$ is also determined by $X^{old}$ due to the fact that the teacher model is trained on $X^{old}$.

First, we defined $\tilde{\textit{Effect}_{old}}$ as the difference between the logits when the old data $x^{old}$ exists or does not exist:
\begin{align}
    &\tilde{\textit{Effect}_{old}} \nonumber\\
    &= \mathbb{P}(Z_{[old]}^{new}|do(X^{old}=x^{old})) - \mathbb{P}(Z_{[old]}^{new}|do(X^{old}=0)). 
\end{align}
Then, we expand the above equation as follows:
\begin{align}
    & \tilde{\textit{Effect}_{old}} &\nonumber \\
    &= \mathbb{P}(Z_{[old]}^{new}|X^{old}=x^{old}) - \mathbb{P}(Z_{[old]}^{new}|X^{old}=0) &\label{eq:effect_old_line0} \\
    &= \mathbb{P}(Z_{[old]}^{new}|{Z_0}^{new}_{[old]},X^{old}=x^{old})\mathbb{P}({Z_0}^{new}_{[old]}|X^{old}=x^{old})& \nonumber \\
    &\quad- \mathbb{P}(Z_{[old]}^{new}|{Z_0}^{new}_{[old]},X^{old}=0)\mathbb{P}({Z_0}^{new}_{[old]}|X^{old}=0)& \label{eq:effect_old_line1} \\
    &= \mathbb{P}(Z_{[old]}^{new}|{Z_0}^{new}_{[old]})\mathbb{P}({Z_0}^{new}_{[old]}|X^{old}=x^{old}) \nonumber \\
    &\quad- \mathbb{P}(Z^{new}_{[old]}|{Z_0}^{new}_{[old]})\mathbb{P}({Z_0}^{new}_{[old]}|X^{old}=0)& \label{eq:effect_old_line2} \\
    &= \mathbb{P}(Z_{[old]}^{new}|{Z_0}^{new}_{[old]}) (\mathbb{P}({Z_0}^{new}_{[old]}|X^{old}=x^{old})& \nonumber \\
    &\quad- \mathbb{P}({Z_0}^{new}_{[old]}|X^{old}=0)).& \label{eq:effect_old_line3}
\end{align}
The Equation \ref{eq:effect_old_line0} holds because $X^{old}$ has no parent nodes.
The Equation \ref{eq:effect_old_line1} holds because of the Bayes rule.
The Equation \ref{eq:effect_old_line2} holds because ${Z_0}^{new}_{[old]}$ is the only mediator \cite{pearl2009causality} from $X^{old}$ to $Z_{[old]}^{new}$.
The Equation \ref{eq:effect_old_line3} shows that $\tilde{\textit{Effect}_{old}}$ can be decomposed to the effect of $\mathbb{P}(Z_{[old]}^{new}|{Z_0}^{new}_{[old]})$ and $\mathbb{P}({Z_0}^{new}_{[old]}|X^{old}=x^{old}$.
The former term represents how $Z_{[old]}^{new}$ is affected by ${Z_0}^{new}_{[old]}$ and implies the constraint of prediction between teacher and students.
The latter term represents the difference of $Z_{[old]}^{new}$ when the teacher model is pretrained on old data $X^{old}=x^{old}$ or not pretrained $X^{old}=0$, and it is non-zero.

For implementation, we define $\textit{Effect}_{old}$ as the Kullback-Leibler Divergence between teacher and student prediction:
\begin{equation}
    \textit{Effect}_{old} = \mathbb{E}_{(x,y) \sim \mathcal{D}_t}(-\alpha \mathcal{L}_{KL}(S^{new}_{[old]}(x),{S_0}^{new}_{[old]}(x)).
    \label{eq:effect_old}
\end{equation}
$\mathcal{L}_{KL}(\cdot,\cdot)$ is the Kullback-Leibler Divergence.
$(x,y)$ is sampled from $\mathcal{D}_t$.
$\alpha$ is the scaling hyperparameters.
$S^{new}_{[old]}$ and ${S_0}^{new}_{[old]}$ are the scores of old classes predicted by the student and teacher model.

In Figure \ref{fig:bace_effect_old}, we build the following positive causal paths from $X^{old}$ and $X^{new}$ to $Z_{[old]}^{new}$ for adapting to old classes: $X^{old} \rightarrow H_0^{new} \rightarrow {Z_0}^{new}_{[old]} \rightarrow Z^{new}_{[old]}$, $X^{new} \rightarrow H_0^{new} \rightarrow {Z_0}^{new}_{[old]} \rightarrow Z^{new}_{[old]}$ and $X^{new} \rightarrow H^{new} \rightarrow Z_{[old]}^{new}$.
In other words, optimizing $\textit{Effect}_{old}$ encourage both $X^{old}$ and $X^{new}$ to produce positive causal effects on $Z_{[old]}^{new}$ because the teacher model is trained on $X^{old}$ and the prediction relies on $X^{new}$.

In addition, we follow DER++ \cite{buzzega2020dark} to enhance the knowledge distillation process:
\begin{align}
    \textit{Effect}_{old} = & \mathbb{E}_{(x,y) \sim \mathcal{D}_t}(-\alpha \mathcal{L}_{KL}(S^{new}_{[old]}(x),{S_0}^{new}_{[old]}(x)) \nonumber \\
    &+ \mathbb{E}_{(x,y) \sim \mathcal{M}}(-\mathcal{L}_{CE}(x,y) \nonumber \\
    &- ||Z^{buf}_{[old]}(x)-{Z_0}^{buf}_{[old]}(x)||_2^2).
    \label{eq:effect_old_replay}
\end{align}
$\mathcal{L}_{CE}(\cdot,\cdot)$ is the cross-entropy loss.
${Z}^{new}_{[old]}$ and ${Z_0}^{new}_{[old]}$ are the logits of old classes in student and teacher models.
$||\cdot||_2$ is the Euclidean distance.
We note that the proposed causal graphs in Figure \ref{fig:bace_effect_old} are independent of data replay since we consider $X^{old}$ and $X^{buf}$ separately.
Furthermore, we highlight that the knowledge distillation term in Equation \ref{eq:effect_old} is crucial from the causal perspective and should not be discarded as in DER++.

\subsection{$\textit{\textbf{Effect}}_{\textbf{new}}$: Learning New Classes with Balanced Causal Effects from $X^{new}$ and $X^{old}$.}
\label{sec:bace_effect_new}

We propose to learning new classes with positive and balanced causal effects from $X^{old}$ and $X^{new}$.
We denote the prediction score over categories as $S$, which is obtained from logits through the softmax function. 
Firstly, $\textit{Effect}_{new}$ can be defined as the difference between the score prediction of $X^{new}$ when $X^{old}$ exists or does not exist:
\begin{align}
    \textit{Effect}_{new} = & \mathbb{P}(S^{new}|H_0^{new},do(X^{old}=x^{old})) \nonumber \\
    &- \mathbb{P}(S^{new}|H_0^{new},do(X^{old}=0)), 
    \label{eq:effect_new_all}
\end{align}
where $do(\cdot)$ is the \textit{do-operation}, which represents assigning a certain value to a variable without considering its parent nodes.
$\mathbb{P}(S^{new}|H_0^{new},do(X^{old}=x^{old}))$ is the score prediction when $f^{t-1}$ is trained on $x^{old}$.
$\mathbb{P}(S^{new}|H_0^{new}, do(X^{old}=0))$ is the score prediction when the teacher model is trained without old data.
In other words, the teacher model is randomly-initialized. 

Then, we rewrite $\textit{Effect}_{new}$ as the sum of the causal effect on each sample's prediction:
\begin{equation}
    \textit{Effect}_{new} = \sum_i^{N}\textit{Effect}_{new}^{(i)}. \label{eq:effect_new_i_line1}
\end{equation}
By introducing the definition in Equation \ref{eq:effect_new_all}, we have:
\begin{align}
    \textit{Effect}_{new}^{(i)}&= \mathbb{P}({S^{new}}^{(i)}|H_0^{new}=h_0^{(i)},do(X^{old}=x^{old})) \nonumber \\
    &\quad- \mathbb{P}({S^{new}}^{(i)}|H_0^{new}=h_0^{(i)},do(X^{old}=0)) \label{eq:effect_new_i_line2} \\
    &= \mathbb{P}({S^{new}}^{(i)}|H_0^{new}=h_0^{(i)},X^{old}=x^{old}) \nonumber \\
    &\quad- \mathbb{P}({S^{new}}^{(i)}|H_0^{new}=h_0^{(i)},X^{old}=0),
    \label{eq:effect_new_i_line3}
\end{align}
where ${S^{new}}^{(i)}$ is the score prediction of the $i$-th sample $x^{(i)}$.
$h_0^{(i)}$ is the feature of $x^{(i)}$ extracted by the encoder of the teacher model (denoted as $f^{t-1}_{enc}$, i.e., $h_0^{(i)}=f_{enc}^{t-1}(x^{(i)})$.
$N$ is the number of samples in $X^{new}$.
Equation \ref{eq:effect_new_i_line2} defines the causal effect on each sample (i.e., $\textit{Effect}_{new}^{(i)}$) as the difference between the score prediction of $x^{(i)}$ when the teacher model is trained on $x^{old}$ or not.
Equation \ref{eq:effect_new_i_line3} holds because $X^{old}$ has no parent nodes.

And Then, $\textit{Effect}_{new}^{(i)}$ is estimated as follows:
\begin{align}
& \nonumber \textit{Effect}_{new}^{(i)} \\
&= \sum_{k=1}^N (\mathbb{P}({S^{new}}^{(i)}|X^{new}=x^{(k)},H_0^{new}=h_0^{(i)}) \nonumber \\
&\quad(\mathbb{P}(X^{new}=x^{(k)}|H_0^{new}=h_0^{(i)},X^{old}=x^{old}) \nonumber \\ 
&\quad-\mathbb{P}(X^{new}=x^{(k)}|H_0^{new}=h_0^{(i)},X^{old}=0))\label{eq:effect_new_i_line4}\\
&=\sum_{k=1}^N (\mathbb{P}({S^{new}}^{(i)}|X^{new}=x^{(k)}) \nonumber\\
&\quad(\mathbb{P}(X^{new}=x^{(k)}|H_0^{new}=h_0^{(i)},X^{old}=x^{old}) \label{eq:effect_new_i_line5} \nonumber \\ 
&\quad-\mathbb{P}(X^{new}=x^{(k)}|H_0^{new}=h_0^{(i)},X^{old}=0) ) \\
&\approx \sum_{k=1}^N \mathbb{P}({S^{new}}^{(i)}|X^{new}=x^{(k)}) \nonumber \\
&\quad\underbrace{\mathbb{P}(X^{new}=x^{(k)}|H_0^{new}=h_0^{(i)},X^{old}=x^{old})}_{W_{i,k}} \label{eq:effect_new_i_line6} \\
&\approx \sum_{k=1}^K \mathbb{P}({S^{new}}^{(i)}|X^{new}=x^{(k)}) W_{i,k} \label{eq:effect_new_i_line7}
\end{align}
The Equation \ref{eq:effect_new_i_line4} is obtained by applying the Bayes rule to Equation \ref{eq:effect_new_i_line3}.
The Equation \ref{eq:effect_new_i_line5} holds since $X^{new}$ is the only mediator \cite{pearl2009causality} from $X^{old}$ to ${S^{new}}^{(i)}$.
The Equation \ref{eq:effect_new_i_line6} approximates $\mathbb{P}(X^{new}=x^{(k)}|H_0^{new}=h_0^{(i)},X^{old}=0)$ as zero because the likelihood is small when the teacher model is randomly initialized.
Using the Bayes rule, the latter term of Equation \ref{eq:effect_new_i_line6} is proportional to the likelihood term as follows:
\begin{align}
   & \nonumber \mathbb{P}(X^{new}=x^{(k)}|H_0^{new}=h_0^{(i)},X^{old}=x^{old}) \propto \\
   &\mathbb{P}(H_0^{new}=h_0^{(i)}|X^{new}=x^{(k)},X^{old}=x^{old}),
   \label{eq:effect_new_i_line8}
\end{align}
where the likelihood term represents how likely the hidden feature
is $h_0^{(i)}$ when the input sample is $x^{(k)}$.
Obviously, the maximum likelihood is reached when $k=i$ and becomes smaller as the distance between the hidden feature of
$x^{(k)}$ and $h_0^{(i)}$ increases.
Recall that $h_0^{(i)}$ is the hidden feature of $x^{(i)}$ extracted by the teacher model.
The latter term in Equation \ref{eq:effect_new_i_line6} can be regarded as a scaling factor, which is determined by the distance between $x^{(i)}$ and $x^{(k)}$ in the feature space of teacher models.
Considering that it is prohibitive to estimate Equation \ref{eq:effect_new_i_line6} on all training samples due to time and space limitations, we truncate the top-K samples and obtain Equation \ref{eq:effect_new_i_line7}.
In other words, Equation \ref{eq:effect_new_i_line7} computes $\textit{Effect}_{new}^{(i)}$ as the weighted sum of  $\mathbb{P}({S^{new}}^{(i)}|X^{new}=x^{(k)})$ on the K-Nearest-Neighbours of $x^{(i)}$.
And the weight $W_{i,k}$ is larger when the distance between $x^{(k)}$ and $x^{(i)}$ in the feature space of the teacher model gets smaller.

We notice that, when $k=i$, $\mathbb{P}({S^{new}}^{(i)}=y^{(i)}|X^{new}=x^{(i)})$ is exactly the likelihood we expected to maximize.
Therefore, maximizing $\textit{Effect}_{new}$ amounts to minimizing the classification loss of each sample, except that the score is the joint score estimated by itself and its neighbours:
\begin{equation}
    \textit{Effect}_{new} = - \mathbb{E}_{(x,y) \sim \mathcal{D}_t} \mathcal{L}_{CE}(\sum_{\tilde{x} \in \{x\} \cup \mathcal{N}(x)} W(\tilde{x},x)S(\tilde{x}),y),
    \label{eq:effect_new}
\end{equation}
where $\sum_{\tilde{x} \in x \cup \mathcal{N}(x)}W(\tilde{x},x)=1$.
$\mathcal{N}$ is the set of K-Nearest-Neighbors (KNNs) in the feature space of teacher models.
$S(\tilde{x})$ is the score prediction of $\tilde{x}$.
$W(\tilde{x},x)$ is the weight of $S(\tilde{x})$ and is defined as follows:
\begin{equation}
W(\tilde{x},x) =  
\begin{cases}
&W_0, \quad \text{when}\quad \tilde{x}=x; \\
&\frac{(1-W_0)/||H_0(\tilde{x})-H_0(x)||_2}{\sum_{\tilde{x}' \in \mathcal{N}(x)}{1/||H_0(\tilde{x}')-H_0(x)||_2}}, \quad\text{otherwise}. 
\end{cases}
\label{eq:knn_weight}
\end{equation}
In Equation \ref{eq:knn_weight},  the weights of neighbours are proportional to the reciprocal of its Euclidean distance to the input sample.
When  $W_0=1$, $\textit{Effect}_{new}$ degenerates to the cross-entropy loss.
In summary, as shown in Figure \ref{fig:bace_effect_new}, we build causal paths from both $X^{old}$ and $X^{new}$ to $Z^{new}_{[new]}$ by optimizing Equation \ref{eq:effect_new}.

\subsection{Connection between BaCE and Existing Methods}

We summarize the algorithm of BaCE in the Appendix.
By the detailed derivation of $\textit{Effect}_{old}$ and $\textit{Effect}_{new}$ in Section \ref{sec:bace_effect_old} and \ref{sec:bace_effect_new}, we achieve the goal of learning each class with positive and balanced causal effects from both $X^{old}$ and $X^{new}$. 
The final forms of $\textit{Effect}_{old}$ and $\textit{Effect}_{new}$ are closely related to the existing technique for CIL, and the connection is shown as follows:
$\textit{Effect}_{old}$ combines the knowledge distillation term in LWF \cite{li2017learning} and and the regularization term in DER++ \cite{buzzega2020dark}.
$\textit{Effect}_{new}$ is built upon DDE \cite{hu2021distilling} and further estimates the weight of neighbours based on the Euclidean distance to input samples.

Although each component in BaCE relies on existing techniques, we prove why each component is necessary from the causal perspective.
In the experiment, we will show that $\textit{Effect}_{old}$ and $\textit{Effect}_{new}$ encourage models to learn from new and old data in a mutually beneficial way and mitigate catastrophic forgetting in CIL.

\section{Experiments}

To verify the effectiveness of BaCE, we conduct experiments on three tasks: continual image classification, continual text classification, and continual named entity recognition. 

\subsection{Experimental Settings}

\subsubsection{Datasets}
We use CIFAR-100 \cite{krizhevsky2009learning}, CIFAR-10 \cite{krizhevsky2009learning}, 5-datasets \cite{ebrahimi2020adversarial}, OminiBenchmark \cite{zhang2022benchmarking}, Tiny-ImageNet \cite{deng2009imagenet}, ObjectNet \cite{barbu2019objectnet}, ImageNet-R \cite{hendrycks2021many}, VTAB \cite{zhai2019large} in this paper.
The introduction of the image classification datasets used in the paper is summarized in Table \ref{tab:datasets_ic_summary}.
The detailed introduction is provided in the Appendix.

\begin{table}[!t]
  \centering
  \caption{The introduction of the image classification datasets used in the paper.}
    \resizebox{\linewidth}{!}{
    \begin{tabular}{lcccc}
    \toprule
          & \# classes & \# training samples & \# test samples & Source \\
    \midrule
    OmniBenchmark & 300   & 89,697  & 5,985  & \href{https://github.com/ZhangYuanhan-AI/OmniBenchmark}{Link} \\
    Tiny-ImageNet & 200   & 100,000  & 10,000  & \href{https://paperswithcode.com/dataset/tiny-imagenet}{Link} \\
    ObjectNet & 200   & 26,509  & 6,628  & \href{https://objectnet.dev/}{Link} \\
    ImageNet-R & 200   & 24,000  & 6,000  & \href{https://github.com/hendrycks/imagenet-r}{Link} \\
    CIFAR100 & 100   & 50,000  & 10,000  & \href{https://www.cs.toronto.edu/~kriz/cifar.html}{Link} \\
    VTAB  & 50    & 1,796  & 8,619  & \href{https://google-research.github.io/task_adaptation/}{Link} \\
    SVHN  & 10    & 73,257  & 26,032  & \href{http://ufldl.stanford.edu/housenumbers/}{Link} \\
    Fashion-MNIST & 10    & 60,000  & 10,000  & \href{https://github.com/zalandoresearch/fashion-mnist}{Link} \\
    CIFAR10 & 10    & 50,000  & 10,000  & \href{https://www.cs.toronto.edu/~kriz/cifar.html}{Link} \\
    MNIST & 10    & 50,000  & 10,000  & \href{http://yann.lecun.com/exdb/mnist/}{Link} \\
    Not-MNIST & 10    & 18,265  & 459   & \href{https://www.kaggle.com/datasets/jwjohnson314/notmnist}{Link} \\
    \bottomrule
    \end{tabular}%
    }
  \label{tab:datasets_ic_summary}%
\end{table}%

\subsubsection{Evaluation Metrics}
We report three metrics for continual learning, including average accuracy after learning task $t$ ($\mathcal{A}_t$) \cite{chaudhry2019tiny}, forgetting (${FGT}$, lower is better) \cite{chaudhry2018riemannian}, and forward transfer (${FWD}$, higher is better) \cite{lopez2017gradient} after learning task $t$.
The definition of the evaluation metrics are in the Appendix.

\subsubsection{Training Details}
We use SGD as the optimizer for all methods and backbones.
The batch size is set as 128.
When using ViT-B/16 as the backbone, we train models on each task with a learning rate of 1e-3. 
The input image is resized to 224$\times$224 to match the ViT pretrainingpretraining process.
For each method, we exploit the herding algorithm \cite{rebuffi2017icarl} to select old representative samples.
The implementation is based on PyTorch \cite{paszke2019pytorch}.
All experiments are run on GeForce RTX 3090 GPU.
We report the average result on three independent runs.
We do not use additional data augmentation except for \textit{RandomHorizontalFlip} and \textit{RandomCrop}.

\subsubsection{hyperparameters}
In $\textit{Effect}_{new}$, we set the number of neighbours $K=5$ and the weight $W_0=0.95$.
In $\textit{Effect}_{old}$, we set $\alpha=1$.

\begin{table}[!t]
  \centering
  \caption{The comparison with competitive baselines on split CIFAR-100. All methods use pretrained ViT-B/16 as the backbone. OOM: Out of GPU memory.}
  \resizebox{\linewidth}{!}{
    \begin{tabular}{clcccccc}
        \toprule
    \multirow{2}[4]{*}{\textbf{Buffer Size}} & \multicolumn{1}{c}{\multirow{2}[4]{*}{\textbf{Method }}} & \multicolumn{3}{c}{\textbf{20 step}} & \multicolumn{3}{c}{\textbf{10 steps}} \\
\cmidrule{3-8}          &       & \textbf{$\mathcal{A}_{last}$ (↑)} & \textbf{FGT (↓)} & \textbf{FWT (↑)} & \textbf{$\mathcal{A}_{last}$ (↑)} & \textbf{FGT (↓)} & \textbf{FWT (↑)} \\
    \midrule
    \multirow{14}[4]{*}{100 } & ER    & 34.33{\tiny±0.59} & 68.21{\tiny±1.27} & 40.24{\tiny±0.94} & 46.60{\tiny±1.36} & 57.46{\tiny±1.02} & 47.43{\tiny±1.79} \\
          & BiC \cite{wu2019large} & 35.96{\tiny±0.70} & 65.52{\tiny±0.96} & 42.06{\tiny±1.24} & 47.38{\tiny±0.54} & 55.72{\tiny±0.58} & 48.62{\tiny±0.42} \\
          & LUCIR \cite{hou2019learning} & 38.40{\tiny±0.58} & 63.85{\tiny±0.42} & 42.14{\tiny±1.58} & 54.47{\tiny±1.06} & 40.94{\tiny±2.43} & 57.16{\tiny±0.86} \\
          & PODNET \cite{douillard2020podnet} & 26.19{\tiny±1.53} & 76.72{\tiny±2.14} & 35.68{\tiny±3.64} & 43.47{\tiny±2.81} & 60.73{\tiny±1.39} & 46.35{\tiny±1.14} \\
          & DDE \cite{hu2021distilling} & 36.16{\tiny±0.39} & 64.71{\tiny±0.81} & 39.74{\tiny±1.39} & 47.09{\tiny±0.58} & 55.92{\tiny±0.76} & 46.11{\tiny±1.03} \\
          & DER++ \cite{buzzega2020dark} & 54.98{\tiny±0.23} & 46.33{\tiny±1.14} & 62.93{\tiny±1.03} & 61.70{\tiny±0.18} & 39.97{\tiny±0.43} & 60.23{\tiny±0.49} \\
          & CLSER \cite{arani2022learning} & 57.53{\tiny±0.48} & 42.85{\tiny±1.05} & 64.73{\tiny±2.66} & 57.38{\tiny±1.24} & 43.70{\tiny±0.41} & 61.92{\tiny±0.77} \\
          & FOSTER \cite{wang2022foster} & 30.62{\tiny±1.28} & /     & /     & 50.21{\tiny±0.15} & /     & / \\
          & MEMO \cite{zhou2023model} & 61.45{\tiny±0.49} & 37.87{\tiny±0.69} & 61.28{\tiny±1.10} & 62.77{\tiny±0.70} & 35.76{\tiny±1.44} & 58.81{\tiny±0.48} \\
          & BEEF \cite{wang2023beef} & OOM   & OOM   & OOM   & OOM   & OOM   & OOM \\
\cmidrule{2-8}          & \cellcolor{black!10}BaCE ($W_0=1,\alpha=0$) & \cellcolor{black!10}54.12{\tiny±0.29} & \cellcolor{black!10}46.06{\tiny±1.64} & \cellcolor{black!10}62.77{\tiny±0.52} & \cellcolor{black!10}61.80{\tiny±0.61} & \cellcolor{black!10}38.19{\tiny±1.58} & \cellcolor{black!10}59.28{\tiny±1.23} \\
          & \cellcolor{black!10}BaCE ($W_0=1$) & \cellcolor{black!10}62.85{\tiny±0.48} & \cellcolor{black!10}38.14{\tiny±1.21} & \cellcolor{black!10}67.11{\tiny±0.85} & \cellcolor{black!10}70.06{\tiny±0.39} & \cellcolor{black!10}30.51{\tiny±1.03} & \cellcolor{black!10}63.47{\tiny±1.42} \\
          & \cellcolor{black!10}BaCE ($\alpha=0$) & \cellcolor{black!10}57.15{\tiny±0.62} & \cellcolor{black!10}43.55{\tiny±0.69} & \cellcolor{black!10}63.14{\tiny±1.60} & \cellcolor{black!10}64.28{\tiny±0.54} & \cellcolor{black!10}37.08{\tiny±0.89} & \cellcolor{black!10}61.48{\tiny±0.78} \\
          & \cellcolor{black!10}BaCE (Ours) & \cellcolor{black!10}\textbf{65.88{\tiny±0.34}} & \cellcolor{black!10}\textbf{34.05{\tiny±0.46}} & \cellcolor{black!10}\textbf{69.53{\tiny±0.61}} & \cellcolor{black!10}\textbf{74.81{\tiny±0.45}} & \cellcolor{black!10}\textbf{24.78{\tiny±0.64}} & \cellcolor{black!10}\textbf{67.14{\tiny±0.71}} \\
    \midrule
    \multirow{14}[4]{*}{500 } & ER    & 70.68{\tiny±0.26} & 29.75{\tiny±0.24} & 67.16{\tiny±1.03} & 70.78{\tiny±0.42} & 30.28{\tiny±0.36} & 62.35{\tiny±0.66} \\
          & BiC \cite{wu2019large} & 72.86{\tiny±0.34} & 28.10{\tiny±0.41} & 68.07{\tiny±0.59} & 74.59{\tiny±0.13} & 24.84{\tiny±0.64} & 65.36{\tiny±1.52} \\
          & LUCIR \cite{hou2019learning} & 73.27{\tiny±0.20} & 26.29{\tiny±0.48} & 69.93{\tiny±0.98} & 74.52{\tiny±0.41} & 21.68{\tiny±1.66} & 66.45{\tiny±1.34} \\
          & PODNET \cite{douillard2020podnet} & 31.10{\tiny±2.86} & 71.54{\tiny±5.27} & 41.54{\tiny±0.78} & 48.29{\tiny±2.01} & 55.42{\tiny±3.19} & 50.43{\tiny±2.84} \\
          & DDE \cite{hu2021distilling} & 74.23{\tiny±0.27} & 25.53{\tiny±0.44} & 70.19{\tiny±1.20} & 72.02{\tiny±0.58} & 28.43{\tiny±1.14} & 64.16{\tiny±0.58} \\
          & DER++ \cite{buzzega2020dark} & 75.83{\tiny±0.35} & 23.72{\tiny±0.59} & 74.65{\tiny±0.64} & 75.17{\tiny±0.29} & 25.96{\tiny±0.61} & 66.26{\tiny±0.95} \\
          & CLSER \cite{arani2022learning} & 77.76{\tiny±0.82} & 21.50{\tiny±1.14} & 75.21{\tiny±1.42} & 78.54{\tiny±0.52} & 19.68{\tiny±0.35} & 68.35{\tiny±0.78} \\
          & FOSTER \cite{wang2022foster} & 70.21{\tiny±0.54} & /     & /     & 76.84{\tiny±0.29} & /     & / \\
          & MEMO \cite{zhou2023model} & 75.83{\tiny±0.42} & 22.56{\tiny±0.27} & 63.27{\tiny±0.59} & 78.96{\tiny±0.35} & 17.65{\tiny±0.46} & 61.72{\tiny±0.93} \\
          & BEEF \cite{wang2023beef} & OOM   & OOM   & OOM   & OOM   & OOM   & OOM \\
\cmidrule{2-8}          & \cellcolor{black!10}BaCE ($W_0=1,\alpha=0$) & \cellcolor{black!10}76.03{\tiny±0.37} & \cellcolor{black!10}23.65{\tiny±1.41} & \cellcolor{black!10}74.59{\tiny±0.42} & \cellcolor{black!10}75.45{\tiny±0.31} & \cellcolor{black!10}24.91{\tiny±0.35} & \cellcolor{black!10}67.49{\tiny±0.35} \\
          & \cellcolor{black!10}BaCE ($W_0=1$) & \cellcolor{black!10}80.14{\tiny±0.28} & \cellcolor{black!10}17.34{\tiny±0.48} & \cellcolor{black!10}75.44{\tiny±0.65} & \cellcolor{black!10}82.13{\tiny±0.22} & \cellcolor{black!10}17.82{\tiny±0.51} & \cellcolor{black!10}69.14{\tiny±0.21} \\
          & \cellcolor{black!10}BaCE ($\alpha=0$) & \cellcolor{black!10}78.54{\tiny±0.13} & \cellcolor{black!10}20.64{\tiny±0.54} & \cellcolor{black!10}74.12{\tiny±0.48} & \cellcolor{black!10}78.60{\tiny±0.25} & \cellcolor{black!10}20.87{\tiny±0.38} & \cellcolor{black!10}68.77{\tiny±0.79} \\
          & \cellcolor{black!10}BaCE (Ours) & \cellcolor{black!10}\textbf{82.46{\tiny±0.25}} & \cellcolor{black!10}\textbf{15.78{\tiny±0.63}} & \cellcolor{black!10}\textbf{76.51{\tiny±0.66}} & \cellcolor{black!10}\textbf{84.59{\tiny±0.20}} & \cellcolor{black!10}\textbf{12.58{\tiny±0.33}} & \cellcolor{black!10}\textbf{70.56{\tiny±0.48}} \\
    \midrule
    $\infty$ & MTL   & 91.67{\tiny±0.16} & \textbf{/} & \textbf{/} & 91.25{\tiny±0.13} & \textbf{/} & \textbf{/} \\
    \bottomrule
    \end{tabular}%
    }
  \label{tab:main_cic_vit_cifar100}%
\end{table}%

\begin{table}[htbp]
\centering
\caption{The comparison with competitive baselines on 5-datasets. All methods use pretrained ViT-B/16 as the backbone.}
    \tiny
    \resizebox{0.7\linewidth}{!}{
    \begin{tabular}{clcc}
        \toprule
    \textbf{Buffer Size} & \multicolumn{1}{c}{\textbf{Method}} & \textbf{$\mathcal{A}_{last}$ (↑)} & \textbf{FGT (↓)} \\
    \midrule
    \multirow{14}[4]{*}{\textbf{250 }} & ER $\dagger$ & 80.32{\tiny±0.55} & 15.69{\tiny±0.89} \\
          & Gdumb $\dagger$ \cite{prabhu2020gdumb} & 56.99{\tiny±0.06} & / \\
          & BiC $\dagger$ \cite{wu2019large} & 78.74{\tiny±1.41} & 21.15{\tiny±1.00} \\
          & DER++ $\dagger$ \cite{buzzega2020dark} & 80.81{\tiny±0.07} & 14.38{\tiny±0.35} \\
          & Co2L $\dagger$ \cite{cha2021co2l} & 82.25{\tiny±1.17} & 17.52{\tiny±1.35} \\
          & L2P $\dagger$ \cite{wang2022learning} & 85.56{\tiny±0.95} & \textbf{4.22{\tiny±0.03}} \\
          & CLSER \cite{arani2022learning} & 86.51{\tiny±0.36} & 10.58{\tiny±0.57} \\
          & FOSTER \cite{wang2022foster} & 74.21{\tiny±0.17} & / \\
          & MEMO \cite{zhou2023model} & 86.53{\tiny±0.31} & 9.09{\tiny±0.36} \\
          & BEEF \cite{wang2023beef} & 78.09{\tiny±0.22} & / \\
\cmidrule{2-4}          & \cellcolor{black!10}BaCE ($W_0=1,\alpha=0$) & \cellcolor{black!10}84.36{\tiny±0.62} & \cellcolor{black!10}9.58{\tiny±1.56} \\
          & \cellcolor{black!10}BaCE ($W_0=1$) & \cellcolor{black!10}86.10{\tiny±0.86} & \cellcolor{black!10}9.16{\tiny±0.97} \\
          & \cellcolor{black!10}BaCE ($\alpha=0$) & \cellcolor{black!10}86.69{\tiny±0.40} & \cellcolor{black!10}10.58{\tiny±1.13} \\
          & \cellcolor{black!10}BaCE (Ours) & \cellcolor{black!10}\textbf{87.50{\tiny±0.57}} & \cellcolor{black!10}8.41{\tiny±1.27} \\
    \midrule
    \multirow{14}[4]{*}{\textbf{500 }} & ER $\dagger$ & 84.26{\tiny±0.84} & 12.85{\tiny±0.62} \\
          & Gdumb $\dagger$ \cite{prabhu2020gdumb} & 70.76{\tiny±0.12} & / \\
          & BiC $\dagger$ \cite{wu2019large} & 85.53{\tiny±2.06} & 10.27{\tiny±1.32} \\
          & DER++ $\dagger$ \cite{buzzega2020dark} & 84.88{\tiny±0.57} & 10.46{\tiny±1.02} \\
          & Co2L $\dagger$ \cite{cha2021co2l} & 86.05{\tiny±1.03} & 12.28{\tiny±1.44} \\
          & L2P $\dagger$ \cite{wang2022learning} & 88.95{\tiny±0.78} & \textbf{4.92{\tiny±0.71}} \\
          & CLSER \cite{arani2022learning} & 89.43{\tiny±0.26} & 6.20{\tiny±0.75} \\
          & FOSTER \cite{wang2022foster} & 74.96{\tiny±0.30} & / \\
          & MEMO \cite{zhou2023model} & 89.59{\tiny±0.15} & 5.37{\tiny±0.24} \\
          & BEEF \cite{wang2023beef} & 79.13{\tiny±0.07} & / \\
\cmidrule{2-4}          & \cellcolor{black!10}BaCE ($W_0=1,\alpha=0$) & \cellcolor{black!10}86.20{\tiny±0.42} & \cellcolor{black!10}9.16{\tiny±1.32} \\
          & \cellcolor{black!10}BaCE ($W_0=1$) & \cellcolor{black!10}88.58{\tiny±0.63} & \cellcolor{black!10}5.64{\tiny±0.48} \\
          & \cellcolor{black!10}BaCE ($\alpha=0$) & \cellcolor{black!10}88.79{\tiny±0.28} & \cellcolor{black!10}6.20{\tiny±0.71} \\
          & \cellcolor{black!10}BaCE (Ours) & \cellcolor{black!10}\textbf{89.80{\tiny±0.27}} & \cellcolor{black!10}5.22{\tiny±0.76} \\
    \midrule
    $\infty$ & MTL $\dagger$ & 93.93{\tiny±0.18} & \textbf{/} \\
    \bottomrule
    \end{tabular}%
    }
  \label{tab:main_cic_vit_5datasets}%
\end{table}

\subsubsection{Baselines.} 
We consider the following competitive CIL methods: Experience Replay (ER), LwF \cite{li2017learning}, EWC \cite{kirkpatrick2017overcoming}, BiC \cite{wu2019large}, LUCIR \cite{hou2019learning}, PODNET \cite{douillard2020podnet}, DDE \cite{hu2021distilling}, DER++ \cite{buzzega2020dark}, L2P \cite{wang2022learning}, Co2L \cite{cha2021co2l}, Gdumb \cite{prabhu2020gdumb}, CLSER \cite{arani2022learning}, FOSTER \cite{wang2022foster}, MEMO\cite{zhou2023model}, BEEF \cite{wang2023beef}.
Sequential Training (SEQ) and Multi-Task Learning (MTL) are the lower and upper limits of CIL.
We load the same backbone model for all baselines and our method.
The detailed introduction is provided in the Appendix.

\subsection{Results and Analysis}

\subsubsection{Comparison with Baselines} 
The comparison between BaCE and various CIL baselines is shown in Table
\ref{tab:main_cic_vit_cifar100} and \ref{tab:main_cic_vit_5datasets}.
All methods use ViT-B/16 as the backbone.
``OOM'' refers to Out-Of-GPU memory.
Table
\ref{tab:main_cic_vit_cifar100} and \ref{tab:main_cic_vit_5datasets} shows that BaCE performs better than a series of competitive CIL methods.
The result also shows that BaCE has a lower FGT and a higher FWT compared with other methods, indicating that balancing the causal effects is beneficial for preserving old knowledge and learning new concepts.

\subsubsection{Ablation Study}

We consider three ablated versions of BaCE:
(1) BaCE ($W_0=1$): $\textit{Effect}_{new}$ degenerates as the traditional cross-entropy loss on new data. 
From the causal perspective, the causal path $X^{old} \leftrightarrow H_0^{new} \leftrightarrow X^{new}$ is removed in Figure \ref{fig:bace_effect_new}.
(2) BaCE ($\alpha=0$): the distillation term on new data in $\textit{Effect}_{old}$ is removed.
From the causal perspective, all causal paths in Figure \ref{fig:bace_effect_old} are removed.
(3) BaCE ($W_0=1,\alpha=0$): combine (1) and (2).

$\textit{Effect}_{old}$ inflates performance by exploiting the causal effects of new data, which is overlooked in previous works \cite{buzzega2020dark}.
In addition, $\textit{Effect}_{new}$ brings about considerable improvements in various buffer size settings, suggesting that introducing the effect of old data to learning new classes alleviates the causal imbalance problem and reduces the forgetting of old knowledge.

\subsubsection{Visualization of the KNN Examples in $\textit{Effect}_{new}$} 

\begin{figure}[!t]
    \centering
    \includegraphics[width=0.8\linewidth]{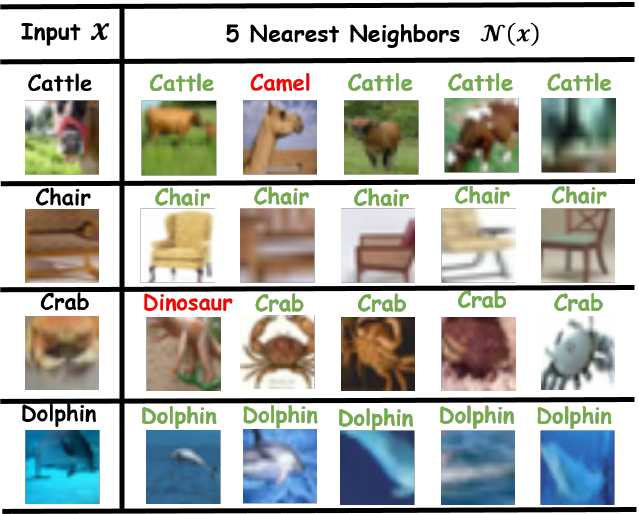}
    \caption{The KNNs in $\textit{Effect}_{new}$ on 5-datasets}
    \label{fig:case_study_1}
\end{figure}

We provide more KNN examples 5-datasets and CIFAR100 in Figure \ref{fig:case_study_1} and the Appendix. 
The ground-truth label is on the top of each image. 
The {\color{teal} green} and {\color{red} red} labels of neighbours represent whether or not they are the same as the input sample.
Figure \ref{fig:case_study_1} shows that the teacher model can recognize new classes with prior knowledge to some extent.
It is an intuitive way to understand how $\textit{Effect}_{new}$ consider the causal effects of both new and old data.
Although some neighbours may have different categories from those of input samples, input samples and their neighbours bear a resemblance in the feature space of the teacher model and thus share the same prior knowledge about input samples.
Therefore, optimizing joint scores encourages models to preserve prior knowledge when adapting to new classes.

\subsubsection{Hyper-parameter Analysis}

\begin{figure}[!t]
    \centering
    \subfloat[20 steps]{
    \includegraphics[width=0.45\linewidth]{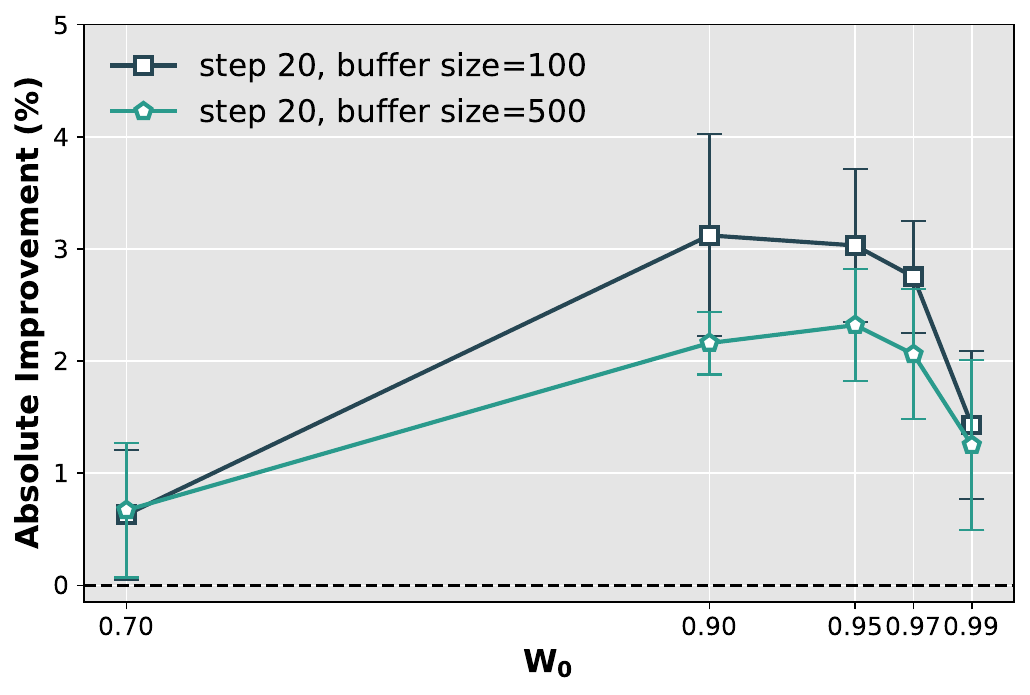}
    }
    \subfloat[10 steps]{
        \includegraphics[width=0.45\linewidth]{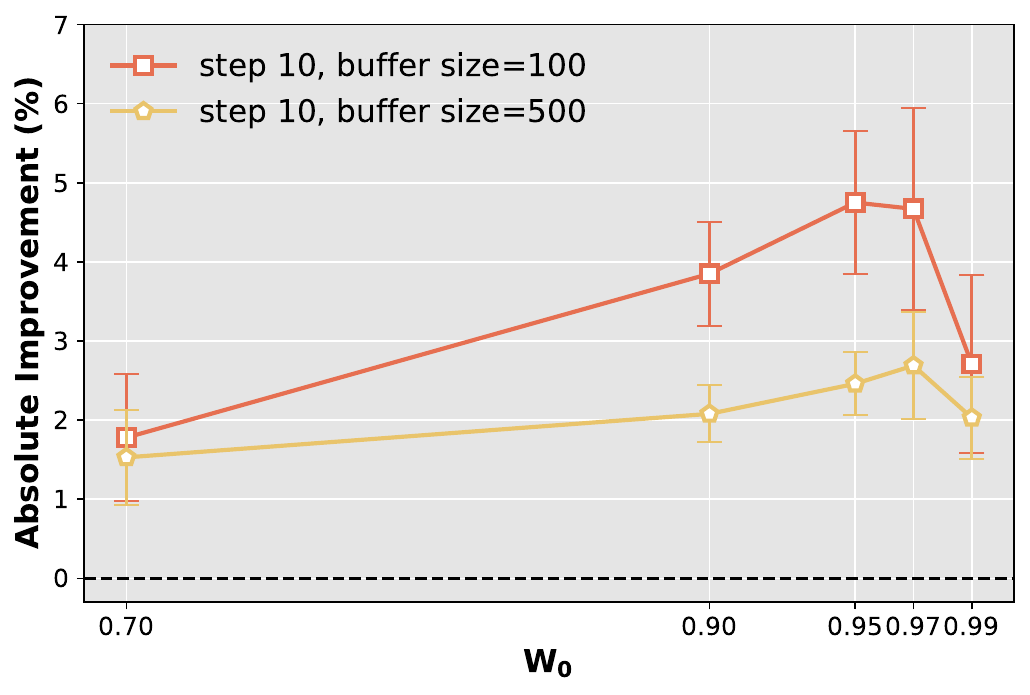}
    }
    \caption{The absolute improvements of $\textit{Effect}_{new}$.}
    \label{fig:hyper-parameter-new}
\end{figure}

\begin{table}[!t]
  \caption{The hyper-parameter analysis of $\textit{Effect}_{old}$.}
  \resizebox{\linewidth}{!}{
    \begin{tabular}{clcccc}
    \toprule
    \multirow{2}[4]{*}{\textbf{Buffer Size}} & \multicolumn{1}{c}{\multirow{2}[4]{*}{\textbf{Method}}} & \multicolumn{2}{c}{\textbf{20 Steps}} & \multicolumn{2}{c}{\textbf{10 Steps}} \\
\cmidrule{3-6}          &       & \textbf{$\mathcal{A}_{last}$ (↑)} & \boldmath{}\textbf{$\Delta$}\unboldmath{} & \textbf{$\mathcal{A}_{last}$ (↑)} & \boldmath{}\textbf{$\Delta$}\unboldmath{} \\
    \midrule
    \multirow{4}[2]{*}{100 } & BaCE ($\alpha=0$) & 57.15  & /     & 64.28  & / \\
          & BaCE ($\alpha=0.01$) & 57.68  & +0.53 & 65.21  & +0.93 \\
          & BaCE ($\alpha=0.1$) & 64.13  & +6.98 & 67.73  & +3.45 \\
          & \cellcolor{black!10}BaCE ($\alpha=1$,Ours) & \cellcolor{black!10}\textbf{65.88 } & \cellcolor{black!10}\textbf{+8.73} & \cellcolor{black!10}\textbf{74.81 } & \cellcolor{black!10}\textbf{+10.53} \\
    \midrule
    \multirow{4}[2]{*}{500 } & BaCE ($\alpha=0$) & 78.54  & /     & 78.60  & / \\
          & BaCE ($\alpha=0.01$) & 78.71  & +0.17 & 78.96  & +0.36 \\
          & BaCE ($\alpha=0.1$) & 81.48  & +2.94 & 79.64  & +1.04 \\
          & \cellcolor{black!10}BaCE ($\alpha=1$,Ours) & \cellcolor{black!10}\textbf{82.46 } & \cellcolor{black!10}\textbf{+3.92} & \cellcolor{black!10}\textbf{84.59 } & \cellcolor{black!10}\textbf{+5.99} \\
    \bottomrule
    \end{tabular}%
    \label{tab:hyper-parameter-old}%
    }
\end{table}

\begin{figure}[htbp]
    \centering
    \includegraphics[width=0.7\linewidth]{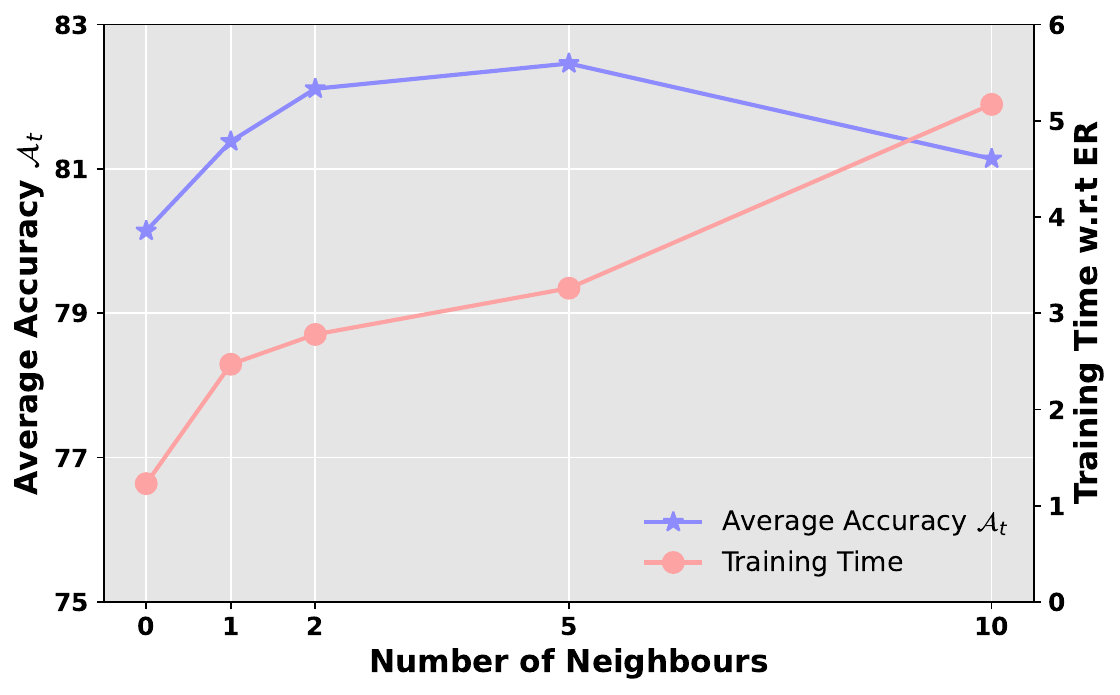}
    \caption{The hyper-parameter analysis of $K$. The dataset is 20-step split CIFAR100. The buffer size is 500.}
    \label{fig:runtime_hyper_analysis}
\end{figure}

The analysis of the hyperparameters in $\textit{Effect}_{new}$ and $\textit{Effect}_{old}$ are summarized in Figure \ref{fig:hyper-parameter-new} and Table \ref{tab:hyper-parameter-old}, respectively.
Figure \ref{fig:hyper-parameter-new} shows the difference between BaCE and BaCE ($W_0=1$) when different $W_0$ is selected. 
It indicates that the model has robust performance when $W_0=0.95$.
Table \ref{tab:hyper-parameter-old} indicates that $\textit{Effect}_{old}$ (Equation \ref{eq:effect_old}) brings considerable improvement based on DER++.

The hyperparameter analysis of $K$ is shown in Figure \ref{fig:runtime_hyper_analysis}.
Figure \ref{fig:runtime_hyper_analysis} shows that BaCE achieves the best performance when $K=5$ and performance is degraded when $K=10$.
It indicates that selecting more neighbours is beneficial when $K<=5$.
However, estimating $\textit{Effect}_{new}$ with a large $K$ may bring noise to the learning process.
In addition, Figure \ref{fig:runtime_hyper_analysis} also indicates that BaCE takes a longer training time than REPLAY.
The reason is that the neighbour relationship and the forward passes of neighbours are computed.
Considering the improvement brought by BaCE, longer training time is acceptable.

\subsubsection{Evaluation on DeiT-S/16 with Less Information Overlapping}

\begin{table}[htbp]
  \centering
  \caption{The average accuracy after the final task. The buffer size for replay-based methods is 2000 for CIFAR100 and TinyImageNet and 200 for CIFAR10. The baseline results are from \cite{DBLP:conf/icml/KimXK023}.}
  \resizebox{\linewidth}{!}{
    \begin{tabular}{lcccccc}
    \toprule
    Method & C10-5T & C100-10T & C100-20T & T-5T  & T-10T & Average \\
    \midrule
    HAT   & 79.36{\tiny±5.16} & 68.99{\tiny±0.21} & 61.83{\tiny±0.62} & 65.85{\tiny±0.60} & 62.05{\tiny±0.55} & 67.62  \\
    OWM   & 41.69{\tiny±6.34} & 21.39{\tiny±3.1}8 & 16.98{\tiny±4.44} & 24.55{\tiny±2.48} & 17.52{\tiny±3.45} & 24.43  \\
    SLDA  & 88.64{\tiny±0.05} & 67.82{\tiny±0.05} & 67.80{\tiny±0.05} & 57.93{\tiny±0.05} & 57.93{\tiny±0.06} & 68.02  \\
    PASS  & 86.21{\tiny±1.10} & 68.90{\tiny±0.94} & 66.77{\tiny±1.18} & 61.03{\tiny±0.38} & 58.34{\tiny±0.42} & 68.25  \\
    L2P   & 73.59{\tiny±4.15} & 61.72{\tiny±0.81} & 53.84{\tiny±1.59} & 59.12{\tiny±0.96} & 54.09{\tiny±1.14} & 60.17  \\
    \midrule
    iCaRL & 87.55{\tiny±0.99} & 68.90{\tiny±0.47} & 69.15{\tiny±0.99} & 53.13{\tiny±1.04} & 51.88{\tiny±2.36} & 66.12  \\
    A-GEM & 56.33{\tiny±7.77} & 25.21{\tiny±4.00} & 21.99{\tiny±4.01} & 30.53{\tiny±3.99} & 21.90{\tiny±5.52} & 36.89  \\
    EEIL  & 82.34{\tiny±3.13} & 68.08{\tiny±0.51} & 63.79{\tiny±0.66} & 53.34{\tiny±0.54} & 50.38{\tiny±0.97} & 63.59  \\
    GD    & 89.16{\tiny±0.53} & 64.36{\tiny±0.57} & 60.10{\tiny±0.74} & 53.01{\tiny±0.97} & 42.48{\tiny±2.53} & 61.82  \\
    DER++ & 84.63{\tiny±2.91} & 69.73{\tiny±0.99} & 70.03{\tiny±1.46} & 55.84{\tiny±2.21} & 54.20{\tiny±3.28} & 66.89  \\
    HAL   & 84.38{\tiny±2.70} & 67.17{\tiny±1.50} & 67.37{\tiny±1.45} & 52.80{\tiny±2.37} & 55.25{\tiny±3.60} & 65.39  \\
    MORE  & 89.16{\tiny±0.96} & 70.23{\tiny±2.27} & 70.53{\tiny±1.09} & 64.97{\tiny±1.28} & 63.06{\tiny±1.26} & 71.59  \\
    ROW   & 90.97{\tiny±0.19} & \textbf{74.72{\tiny±0.48}} & 74.60{\tiny±0.12} & 65.11{\tiny±1.97} & 63.21{\tiny±2.53} & 73.72  \\
    \midrule
    \rowcolor{black!10}BaCE  & \textbf{91.54{\tiny±0.31}} & 74.15{\tiny±0.60} & \textbf{75.06{\tiny±0.71}} & \textbf{66.23{\tiny±1.28}} & \textbf{63.85{\tiny±3.02}} & \textbf{74.02} \\
    \bottomrule
    \end{tabular}%
    }
  \label{tab:main_cic_deit_buf2000}%
\end{table}%

To mitigate information overlapping between pretraining data and the datasets for incremental learning, we follow the experimental setting in \cite{DBLP:conf/icml/KimXK023} to further evaluate the proposed method.
We use DeiT-S/16 \cite{touvron2021training} as the backbone.
We use the same model checkpoint as \cite{DBLP:conf/icml/KimXK023}, which is trained using 611 classes of ImageNet after removing 389 classes similar or identical to the classes of CIFAR and Tiny-ImageNet.
Following the setting in \cite{DBLP:conf/icml/KimXK023}, we finetune adapters \cite{houlsby2019parameter} to leverage the strong performance of the pretrained model while adapting to new knowledge.

For CIFAR10, 10 classes are divided into 5 tasks, with 2 classes for each task (C10-T5).
For CIFAR100, 100 classes are divided into 10 and 20 tasks (C100-T10 and C100-T200.
For TinyImageNet, 200 classes are split into 5 and 10 tasks (T-5T and T-10T).
The result in Table \ref{tab:main_cic_deit_buf2000} indicates that BaCE consistently outperforms the existing CIL methods.

\subsubsection{Evaluation on Challenging Datasets}

\begin{table*}[htbp]
  \centering
  \caption{The average accuracy, the difference of feature embedding distance between new classes and old classes (Feat.Embed.Dist.) and the probing accuracy (Prob.Acc.) after learning the final task on four challenging datasets.}
  \resizebox{\linewidth}{!}{
           \begin{tabular}{lccc|ccc|ccc|ccc}
    \toprule
          & \multicolumn{3}{c}{OmniBenchmark} & \multicolumn{3}{c}{ImageNet-R} & \multicolumn{3}{c}{ObjectNet} & \multicolumn{3}{c}{VTAB} \\
          & Acc.  & Prob.Acc & \multicolumn{1}{c}{Feat.Embed.Dist} & Acc.  & Prob.Acc & \multicolumn{1}{c}{Feat.Embed.Dist} & Acc.  & Prob.Acc & \multicolumn{1}{c}{Feat.Embed.Dist} & Acc.  & Prob.Acc & Feat.Embed.Dist \\
    \midrule
    ER    & 68.50  & 74.51  & 0.1376  & 65.88  & 72.28  & 0.0731  & 46.14  & 52.14  & 0.1485  & 78.53  & 90.37  & 0.1954  \\
    BiC \cite{wu2019large} & 71.84  & 74.36  & 0.0709  & 70.14  & 73.13  & 0.0782  & 48.65  & 52.63  & 0.1377  & 81.65  & 90.15  & 0.1732  \\
    LUCIR \cite{hou2019learning} & 67.18  & 73.85  & 0.1621  & 67.43  & 71.06  & 0.0956  & 47.39  & 52.77  & 0.1365  & 80.49  & 89.79  & 0.1814  \\
    IL2M \cite{belouadah2019il2m} & 69.52  & 74.94  & 0.1230  & 69.39  & 74.70  & 0.0505  & 48.42  & 52.55  & 0.1106  & 80.37  & 90.55  & 0.1623  \\
    DER++ \cite{buzzega2020dark} & 72.11  & 74.68  & 0.1156  & 72.18  & 75.58  & 0.0476  & 49.18  & 52.61  & 0.0974  & 79.09  & 90.62  & 0.1836  \\
    CLSER \cite{arani2022learning} & 72.24  & 74.13  & 0.1235  & 71.93  & 76.02  & 0.0670  & 51.04  & 52.38  & 0.1182  & 79.56  & 90.51  & 0.1878  \\
    \midrule
    BaCE  & 73.30  & 75.17  & 0.0392  & 74.48  & 77.55  & 0.0345  & 52.29  & 53.30  & 0.0825  & 84.86  & 90.84  & 0.0652  \\
    BaCE + Bias Correction & \textbf{73.93 } & \textbf{75.17 } & \textbf{0.0392 } & \textbf{75.20 } & \textbf{77.55 } & \textbf{0.0345 } & \textbf{53.32 } & \textbf{53.30 } & \textbf{0.0825 } & \textbf{85.53 } & \textbf{90.84 } & \textbf{0.0652 } \\
    \bottomrule
    \end{tabular}%
    }
  \label{tab:main_cic_challenge_datasets}%
\end{table*}%

We evaluate BaCE on four challenging datasets that have large domain gaps with ImageNet, namely ImageNet-R \cite{hendrycks2021many}, ObjectNet \cite{barbu2019objectnet}, Omnibenchmark \cite{zhang2022benchmarking} and VTAB \cite{zhai2019large}.
ImageNet-R and ObjectNet contain challenging samples that ImageNet pretrained models cannot handle \cite{zhou2023revisiting}, while Omnibenchmark and VTAB contain diverse classes from multiple complex realms.
We follow the experimental setting in \cite{zhou2023revisiting} and use ViT-B/16-IN21K \cite{dosovitskiy2020image} as the backbone model, which is not supervised-finetuned on ImageNet1K.
We use the datasets sampled by \cite{zhou2023revisiting}, where Omnibenchmark contains 300 classes, ImageNet-R and ObjectNet contain 200 classes, and VTAB contains 50 classes.
We equally divide all classes into 10 tasks for each dataset.

We compare our method with several distillation-based methods: BiC \cite{wu2019large}, IL2M \cite{belouadah2019il2m}, LUCIR \cite{hou2019learning}, DER++ \cite{buzzega2020dark}, CLSER \cite{arani2022learning}.
All baselines and BaCE store 5 samples for each class on ObjectNet, Omnibenchmark, and VTAB and store 20 samples for each class on ImageNet-R.
All baselines and BaCE load the same PTM (i.e., ViT-B/16-IN21K) for training.
We set $\alpha=2$ for BaCE on the four datasets.
The result in Table \ref{tab:main_cic_challenge_datasets} indicates that BaCE outperforms the compared methods on four datasets consistently.

\subsubsection{Comparison with Existing Methods for Class Imbalance Problem}

\begin{figure*}[htbp]
    \centering
    \subfloat[LUCIR]{
        \includegraphics[width=0.18\linewidth]{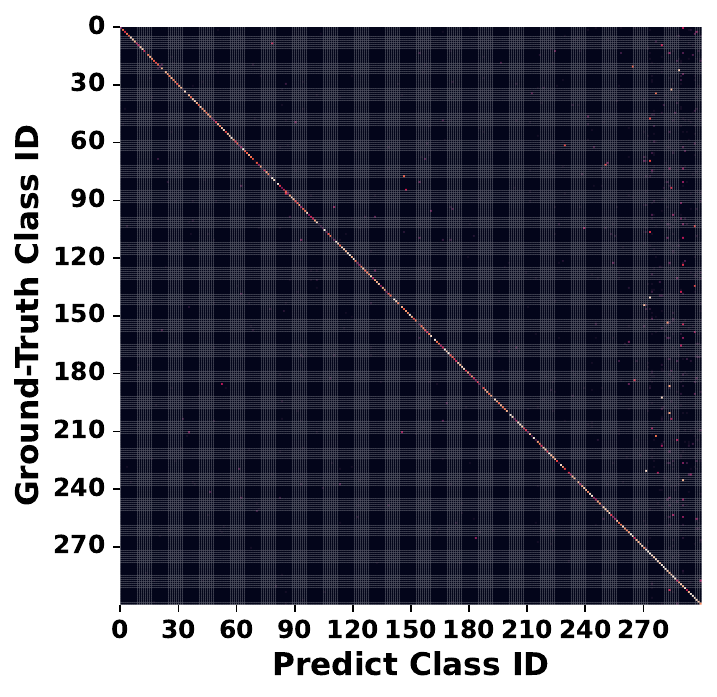}
    }
    \subfloat[IL2M]{
        \includegraphics[width=0.18\linewidth]{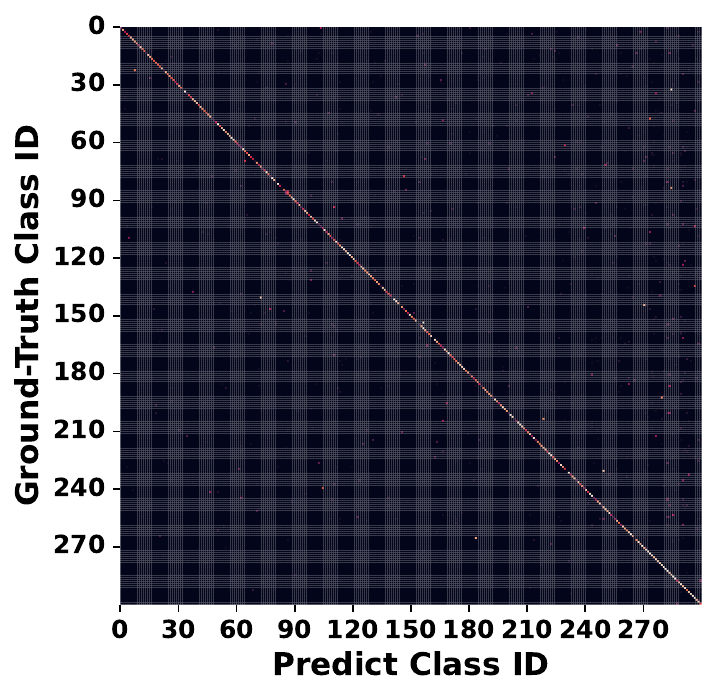}
    }
    \subfloat[BiC]{
        \includegraphics[width=0.18\linewidth]{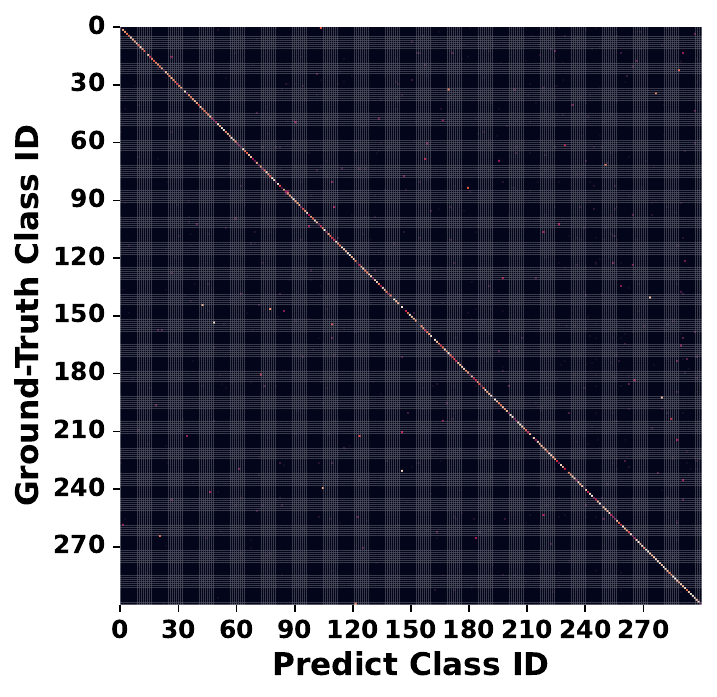}
    }
    \subfloat[BaCE]{
        \includegraphics[width=0.18\linewidth]{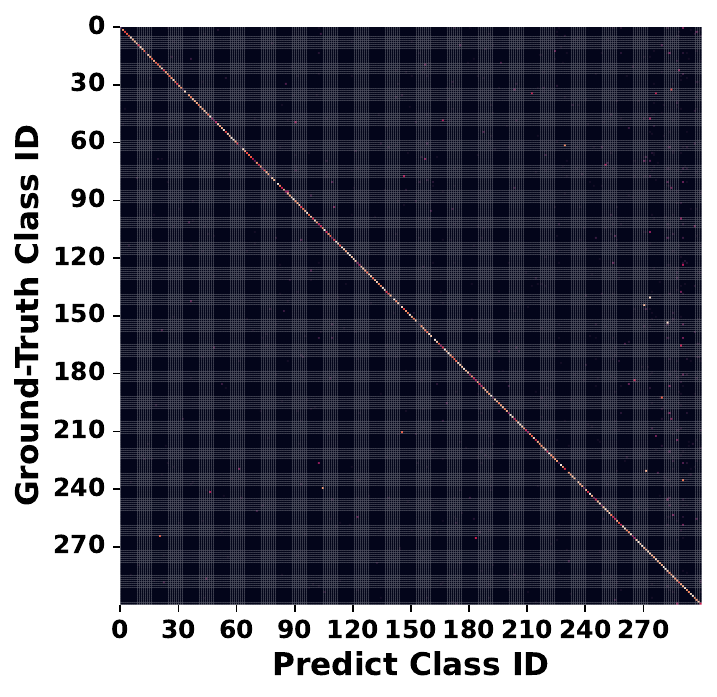}
    }
    \subfloat[BaCE+Bias Correction]{
        \includegraphics[width=0.21\linewidth]{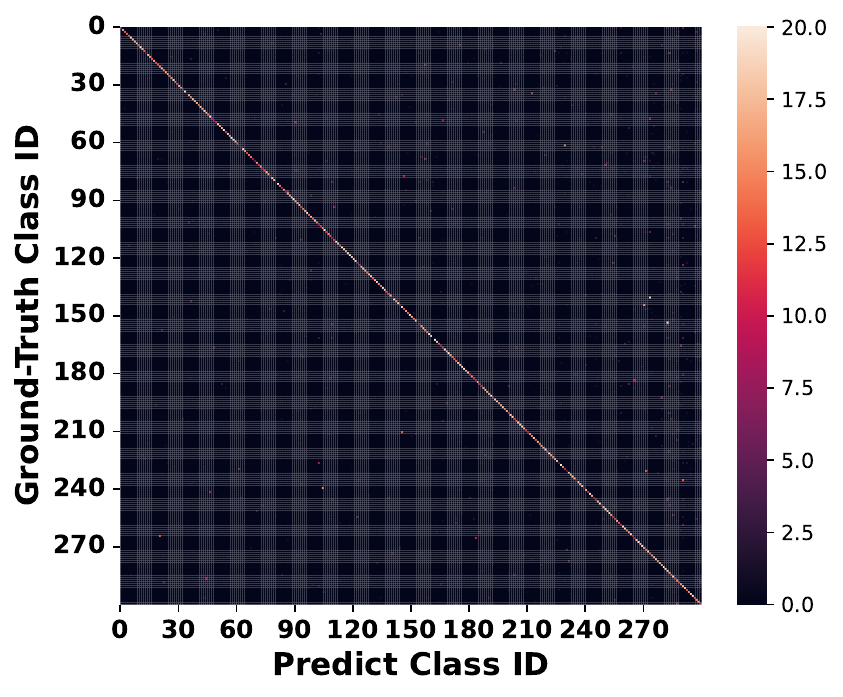}
    }
    \caption{The confusion matrix of different methods on OmniBenchmark.}
    \label{fig:confusion_matrix_OmniBenchmark}
\end{figure*}

We compare BaCE with three classical methods for the class imbalance problem: BiC, LUCIR, and IL2M. 
As shown in Table \ref{tab:main_cic_challenge_datasets}, BaCE reduces the feature embedding distance between new and old classes significantly.
More importantly, it improves the probing accuracy of existing methods designed for the class imbalance problem.
It indicates that balancing the scale of logits alone (e.g., BiC and IL2M) may not necessarily enhance the feature learning process.
In contrast, BaCE pursues causal-balanced learning for each class, thereby achieving higher probing performance in IL.

To have a closer look at the class imbalance problem, we provide the confusion matrices of BaCE, BiC, LUCIR, IL2M on the OmniBenchmark and VTAB challenging datasets in Figure \ref{fig:confusion_matrix_OmniBenchmark} and the Appendix.
BiC shows the most balanced predictions, followed by BaCE and IL2M, and finally, LUCIR.
However, BiC and IL2M make more errors in predicting old classes and perform worse than BaCE.
The reason may be that the old rehearsal data does not reflect the true distribution of the old data, and pursuing a balanced prediction between the rehearsal data and new data diminish the representation ability during IL.

The confusion matrices also show that BiC has a more balanced prediction than BaCE.
Therefore, we rectify the prediction bias towards new classes in BaCE using the re-balancing method in BiC.
Specifically, we follow BiC to learn two additional parameters to correct the bias on the balanced datasets and select the best parameters according to the validation set.
We only re-balance the prediction after learning the last task and the final model is denoted as BaCE+Bias Correction.
The learned bias parameters $(\alpha,\beta)$ on OmniBenchmark and VTAB are $(0.94423,-0.01378)$ and $(0.92134,-0.05441)$, respectively.
As shown in Table \ref{tab:main_cic_challenge_datasets}, correcting the bias slightly improves performance.
Since the encoder and the classifier are unchanged, the probing performance and the feature embedding distance remain unchanged.
The confusion matrices show that ``BaCE+Bias Correction'' achieves a more balanced prediction.

In summary, BaCE pursues training with balanced causal effects, rather than balancing the magnitude of logits like existing methods for the class imbalance problem.
In addition, combining BaCE with existing techniques for the class imbalance problem may further improve the performance.

\subsubsection{Different Variants of $\textit{Effect}_{new}$}

\begin{figure}[htbp]
    \centering
    \subfloat[OmniBenchmark]{
        \includegraphics[width=0.45\linewidth]{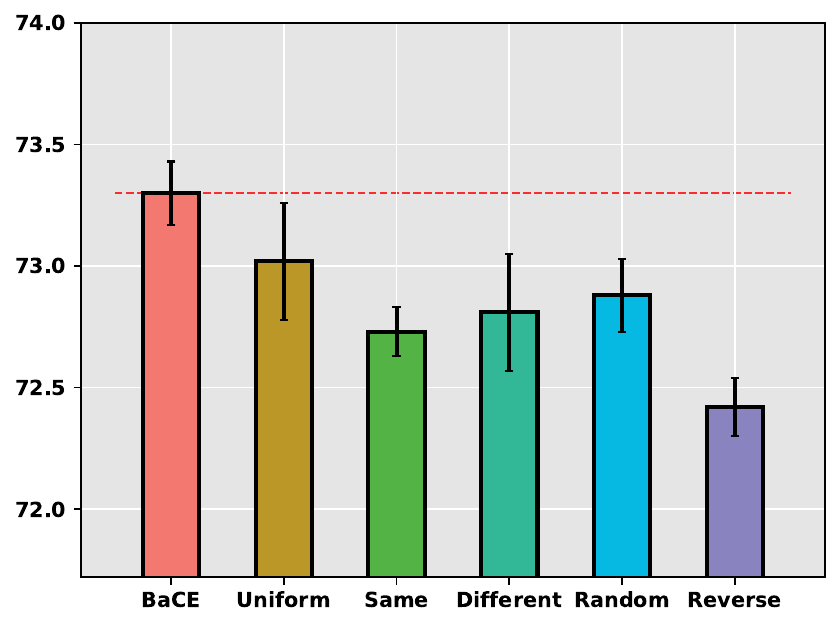}
    }
    \subfloat[ImageNet-R]{
        \includegraphics[width=0.45\linewidth]{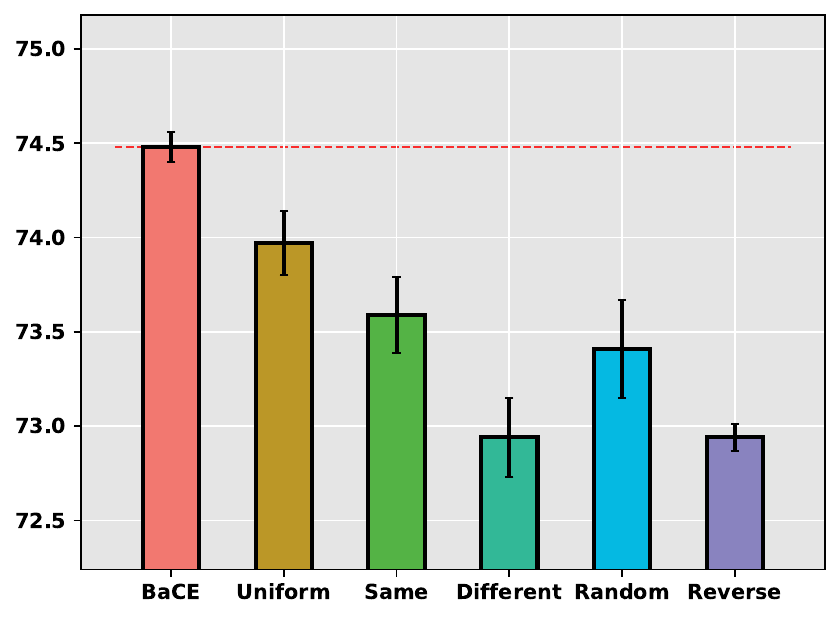}
    }
    \\
    \subfloat[ObjectNet]{
        \includegraphics[width=0.45\linewidth]{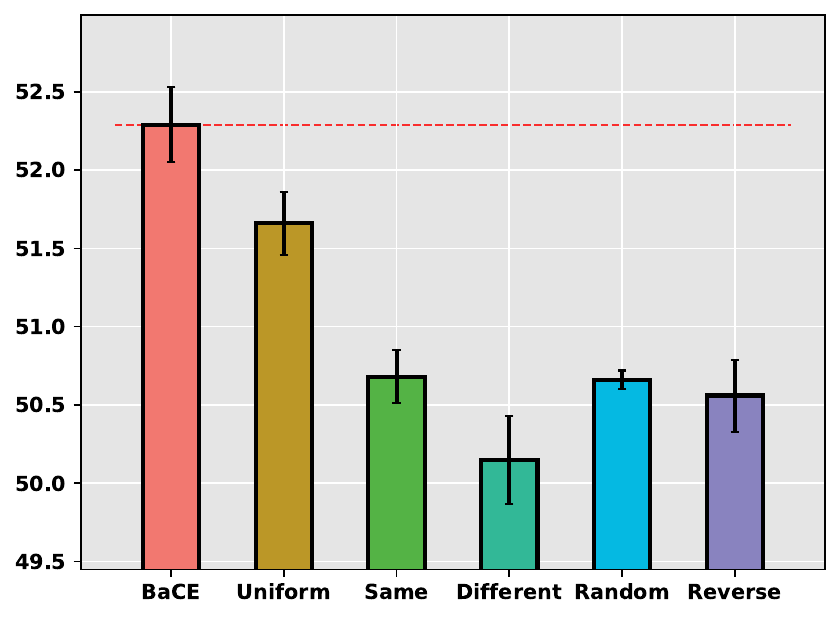}
    }
    \subfloat[VTAB]{
        \includegraphics[width=0.45\linewidth]{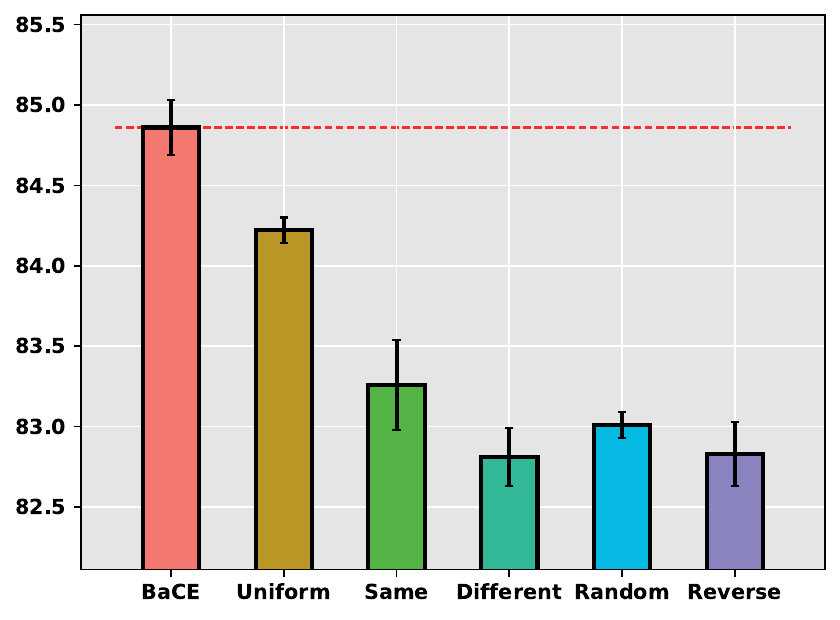}
    }
    \caption{Different variants in $\textit{Effect}_{new}$. The average accuracy after the final task $\mathcal{A}_{last}$ is reported. ``Same'': only the neighbours with the same ground-truth labels as the input sample are selected; ``Different'': only the neighbours with the different ground-truth labels from the input sample are selected; ``Uniform'': the weights of all neighbours are set as the same; ``Random'': the neighbours are randomly sampled from all training data; ``Reverse'': the neighbours with the largest distance are selected.}
    \label{fig:effect_knn_ablation}
\end{figure}

We compare five variants of $\textit{Effect}_{new}$ in Table \ref{fig:effect_knn_ablation}.
The result shows that all ablated variants perform worse than BaCE.
Besides, as shown in Figure \ref{fig:case_study_1}, some neighbours have different labels as the input sample.
It indicates that the teacher model may provide incorrect information about new classes.
However, the performance worsens when we filter out the neighbours with different labels (i.e., ``Same'').
The results show that the KNNs in $\textit{Effect}_{new}$ act as a constraint from the teacher model.
For example, in the first row of Figure \ref{fig:case_study_1}, the teacher model provides the prior knowledge that cattle and camel are animals walking on land.

\section{Conclusion}
In this research, we start from a contradictory phenomenon in recent studies and discover that the conflicting causal effects leads to catastrophic forgetting even when the PTM does not forget. 
To this end, we propose BaCE to mitigate the causal imbalance problem by balancing the causal effects between new and old data when adapting to each class.
Unlike prior CIL methods for the class imbalance problem, BaCE enables models to learn new and old data jointly with the balanced causal effects instead of merely balancing the final prediction.
Finally, we verify the effectiveness of BaCE through extensive experiments on both visual and NLP tasks.

There are two main limitations to this research.
The proposed BaCE requires longer training time than experience replay.
In addition, the performance of BaCE is still unsatisfactory when the buffer size is small.

\bibliography{reference}
\bibliographystyle{IEEEtran}

\appendix

\begin{algorithm}[htbp] 
\caption{Balancing Causal Effects (BaCE)} \label{alg:BaCE}
\KwIn{
$\mathcal{D}_t=\{(x^{(i)},y^{(i)})\}_{i=1}^{N}$: the training set of the task $t$; 
$f^{t-1}$: the teacher model trained on previous $t-1$ tasks;
$K$: the number of neighbours;
$\beta$: the hyper-parameter for controlling the update speed of the teacher model;
$\mathcal{M}$: the rehearsal buffer; 
}
\KwOut{
$f^{t}$: the student model adapted to task $t$
}
Initialize the student model $f^{t} \leftarrow f^{t-1}$\;
\While{\textit{not converge}}{
    Compute the $K$ nearest neighbors of each sample and obtain $\mathcal{N}$ \;
    \For{$(x^{(i)},y^{(i)})$ in $\mathcal{D}_t$}{
        Compute $\textit{Effect}_{new}$\;
        Compute $\textit{Effect}_{old}$\;
        $f^{t} \leftarrow \mathop{\arg\max}\limits_{f^{t}} \textit{Effect}_{new} + \textit{Effect}_{old}$\;
    }
    $f^{t-1} \leftarrow \beta f^{t-1} + (1-\beta) f^t$\;
}
\end{algorithm}

\subsection{Evaluation Metrics}
We report three metrics for continual learning, including average accuracy after learning task $t$ ($\mathcal{A}_t$) \cite{chaudhry2019tiny}, forgetting (${FGT}$, lower is better) \cite{chaudhry2018riemannian}, and forward transfer (${FWD}$, higher is better) \cite{lopez2017gradient} after learning task $t$.
Specifically, 
\begin{equation}
    \mathcal{A}_t = \frac{1}{t}\Sigma_{i=1}^{t}a_{T,i},
\end{equation}
where $a_{t,i}$ is the test accuracy of task $i$ after learning task $t$.
$\mathcal{A}_{last}$ refers to the average accuracy on all learnt tasks after learning the last task.
The forgetting is computed as the average performance degradation of all learned tasks:
\begin{equation}
    FGT = \frac{1}{T-1}\Sigma_{i=1}^{T-1}[ max_{j<t}\{a_{j,i}\} - a_{t,i}],
\end{equation}
where $T$ is the number of all tasks.
The forward transfer is computed as the difference between continual learning and direct instruction-tuning on task $t$: 
\begin{equation}
    FWD = \frac{1}{T-1}\Sigma_{i=2}^{T}[ a_{i-1,i} - \tilde{a_{i}}],
\end{equation}
where $T$ is the number of all tasks.
$\tilde{a_{i}}$ represents the test accuracy for the task $t$ at random initialization.
A larger $FWD$ implies the continual learning algorithm better encourages positive forward transfer for learning task $t$. 

\subsection{Datasets}

We use CIFAR-100 \cite{krizhevsky2009learning}, CIFAR-10 \cite{krizhevsky2009learning}, 5-datasets \cite{ebrahimi2020adversarial}, OminiBenchmark \cite{zhang2022benchmarking}, Tiny-ImageNet \cite{deng2009imagenet}, ObjectNet \cite{barbu2019objectnet}, ImageNet-R \cite{hendrycks2021many}, VTAB \cite{zhai2019large} in this paper.
\begin{itemize}
    \item CIFAR100: CIFAR100 contains 60000 32$\times$32 RGB images of 100 categories. Each class has 500 training images and 100 testing images. We follow \cite{wang2022learning,wu2019large} and split 100 classes evenly into 5, 10, and 20 incremental batches. In our experiment, we learned the 100 classes in ascending order. 
    \item 5-datasets: We consider a CIL setting proposed in \cite{ebrahimi2020adversarial}. 5-datasets consists of five image classification datasets: CIFAR-10, MNIST \cite{lecun1998mnist}, Fashion-MNIST \cite{xiao2017fashion}, SVHN \cite{netzer2011reading}, and notMNIST \cite{bulatov2011notmnist}. In our experiments, we train the model according to the same order as in \cite{wang2022learning}: SVHN, MNIST, CIFAR10, NotMNIST, Fashion-MNIST.Although training each dataset alone is not hard, the sequential training of them is fairly challenging even with ImageNet pretrained models \cite{wang2022learning}, since models are susceptible to forgetting when the tasks are diverse \cite{mehta2021empirical}. 
    \item OminiBenchmark: OminiBenchmark is a large benchmark dataset covering more realms and annotating more images of each realm compared with ImageNet-1k.
    \item Tiny-ImageNet:  Tiny ImageNet has 200 classes, and each class has 500 training images, 50 validation images, and 50 test images. The images are down-sampled to 64 x 64 pixels.
    \item ObjectNet: ObjectNet is a large real-world test set for object recognition with control where object backgrounds, rotations, and imaging viewpoints are random.
    \item ImageNet-R:  ImageNet-R(endition) contains art, cartoons, deviantart, graffiti, embroidery, graphics, origami, paintings, patterns, plastic objects, plush objects, sculptures, sketches, tattoos, toys, and video game renditions of ImageNet classes. ImageNet-R has renditions of 200 ImageNet classes, resulting in 30,000 images.
    \item VTAB: The Visual Task Adaptation Benchmark (VTAB) is a diverse and challenging suite of tasks designed to evaluate general visual representations. VTAB defines a good general visual representation as one that performs well on unseen tasks when trained on limited task-specific data.
\end{itemize}

\begin{table}[htbp]
  \centering
  \caption{The introduction of the visual transformers used in the paper.}
    \begin{tabular}{lcc}
    \toprule
          & \# params & Source \\
    \midrule
    ViT-B/16 & 86M   & \href{https://storage.googleapis.com/vit_models/augreg/B_16-i21k-300ep-lr_0.001-aug_medium1-wd_0.1-do_0.0-sd_0.0--imagenet2012-steps_20k-lr_0.01-res_224.npz}{Link} \\
    ViT-B/16-IN21K & 86M   & \href{https://storage.googleapis.com/vit_models/augreg/B_16-i21k-300ep-lr_0.001-aug_medium1-wd_0.1-do_0.0-sd_0.0.npz}{Link} \\
    DeiT-S/16 & 22M   & \href{https://github.com/k-gyuhak/CLOOD}{Link} \\
    \bottomrule
    \end{tabular}%
  \label{tab:models_ic_summary}%
\end{table}%

\subsection{Backbone Models}
We use ViT-B/16, ViT-B/16-IN21K \cite{dosovitskiy2020image}, DeiT-S/16 \cite{touvron2021training} as the backbone.
The introduction of the backbone models used in the paper is summarized in Table \ref{tab:models_ic_summary}.

\subsection{Baselines} 
We consider the following competitive CIL methods: Experience Replay (ER), LwF \cite{li2017learning}, EWC \cite{kirkpatrick2017overcoming}, BiC \cite{wu2019large}, LUCIR \cite{hou2019learning}, PODNET \cite{douillard2020podnet}, DDE \cite{hu2021distilling}, DER++ \cite{buzzega2020dark}, L2P \cite{wang2022learning}, Co2L \cite{cha2021co2l}, Gdumb \cite{prabhu2020gdumb}, CLSER \cite{arani2022learning}, FOSTER \cite{wang2022foster}, MEMO\cite{zhou2023model}, BEEF \cite{wang2023beef}.
Sequential Training (SEQ) and Multi-Task Learning (MTL) are the lower and upper limits of CIL.
We load the same backbone model for all baselines and our method.
\begin{itemize}
    \item ER \cite{chaudhry2019tiny}: ER is a simple rehearsal method that stores old examples in memory buffers for later replay. Although it is simple, it is an effective and stable baseline.
     \item LwF \cite{li2017learning}: LwF is a method that exploits knowledge distillation, where the teacher is the model learned on previous tasks and the student is the new model. We set the trade-off parameter in LwF as 5 for a fair comparison. 
    \item EWC \cite{kirkpatrick2017overcoming}: EWC is a regularization-based method that slows down the updates on important parameters. We set the weight of the regularization term as 5000.
    \item BiC \cite{hou2019learning}: BiC is an exemplar-based method that adds a regularization item based on knowledge distillation. After each CIL step, BiC finetunes the linear layer and learns two additional parameters to reduce the bias of the backbone network. 
    \item IL2M \cite{belouadah2019il2m}: Compared to exemplar-based methods, IL2M additionally stores old class statistics to reduce the prediction bias towards new classes. 
    \item iCaRL \cite{rebuffi2017icarl}: iCaRL proposes a herding algorithm for selecting old representative samples. In addition, iCaRL adopts a nearest-mean-of-exemplar classifier for classification.
    \item LUCIR \cite{hou2019learning}: LUCIR is an exemplar-based method. LUCIR proposes some modifications to promote separation in the feature space and generate more coordinated incremental learning classifiers. The initial weight of the distillation loss $\lambda_{base}$ is set to 5, $K$ is set to 2, and $m$ is set to 0.5.
    \item PODNet \cite{douillard2020podnet}: Apart from the classification loss of new data, PODNet constrains the output of each intermediate layer and the feature output by the backbone network.
    \item DDE \cite{hu2021distilling}: DDE is based on causal inference, which proposes to extract the causal effect between new and old data and capture the incremental momentum effect of the data flow. 
    \item L2P \cite{wang2022learning}: L2P learns a set of prompts that dynamically instruct models to solve corresponding tasks. The set of prompts is called a prompt pool, which is structured in a key-value shared memory space. 
    \item CLSER \cite{arani2022learning}: CLSER improves the replay buffer system with the complementary learning system (CLS) theory. 
    \item FOSTER \cite{wang2022foster}: FOSTER is a two-stage architecture-based method. The boosting epochs, compression epochs, and epochs of the initial task are set to 5. 
    \item MEMO \cite{zhou2023model}: MEMO is a simple yet effective architecture-based method. MEMO extends specialized layers based on the shared generalized representations. In our experiments, we regard the topmost transformer layer as the specialized layer, and the other transformer layers as the task-agnostic layers.
    \item BEEF \cite{wang2023beef}: BEEF is an architecture-based CIL method based on energy-based theory. BEEF decouples the training of independent modules while achieving bidirectional compatibility among modules. The expansion epoch number and the fusion epoch number are set to 3, and the epoch number of the initial task is set to 5.
\end{itemize}

\subsection{Comparison with Baselines} 
\begin{figure*}[htbp]
    \centering
    \subfloat[20 steps, 100 buffer size]{
        \includegraphics[width=0.49\linewidth]{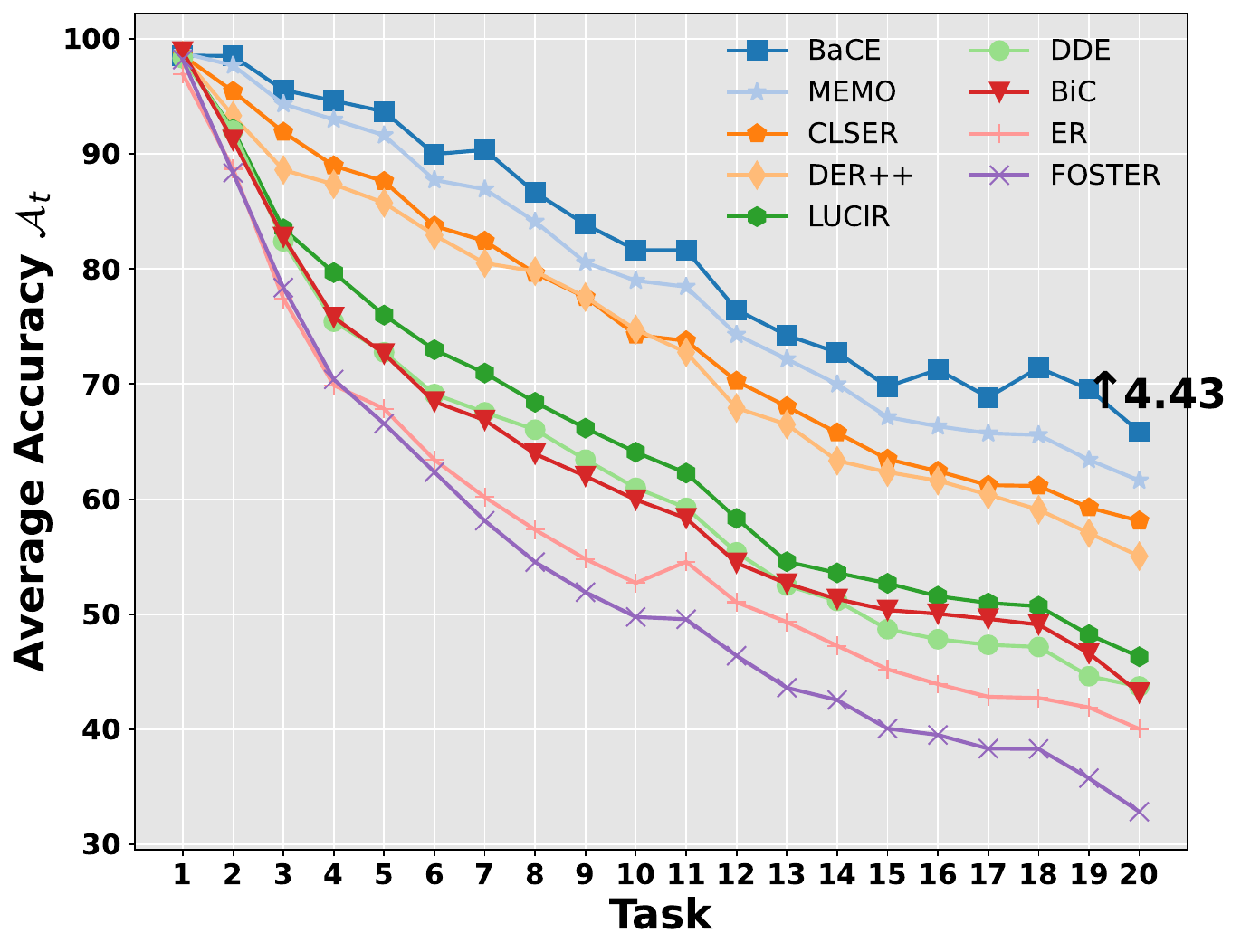}
    }
    \subfloat[20 steps, 500 buffer size]{
        \includegraphics[width=0.49\linewidth]{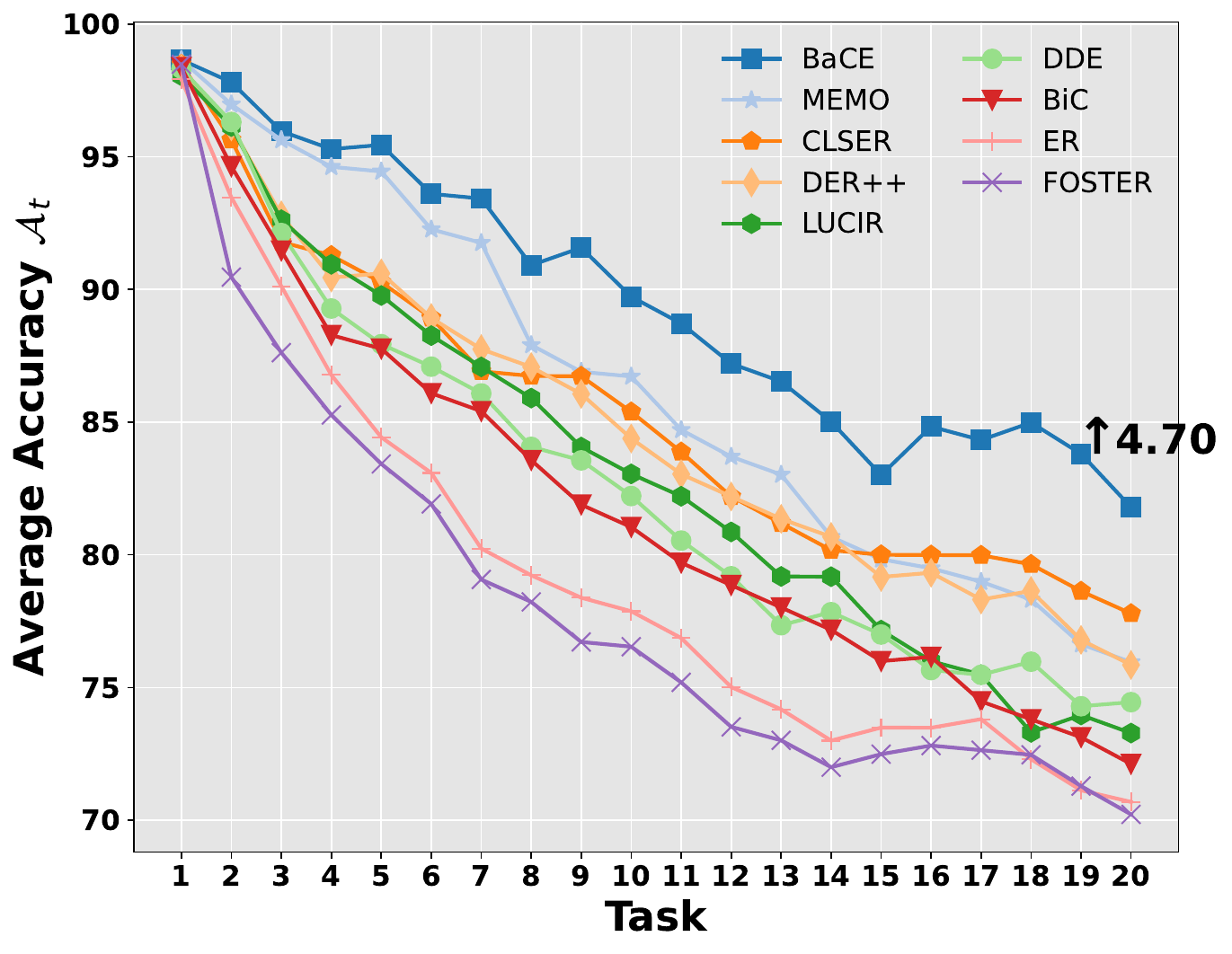}
    }
    \\
    \subfloat[10 steps, 100 buffer size]{
        \includegraphics[width=0.49\linewidth]{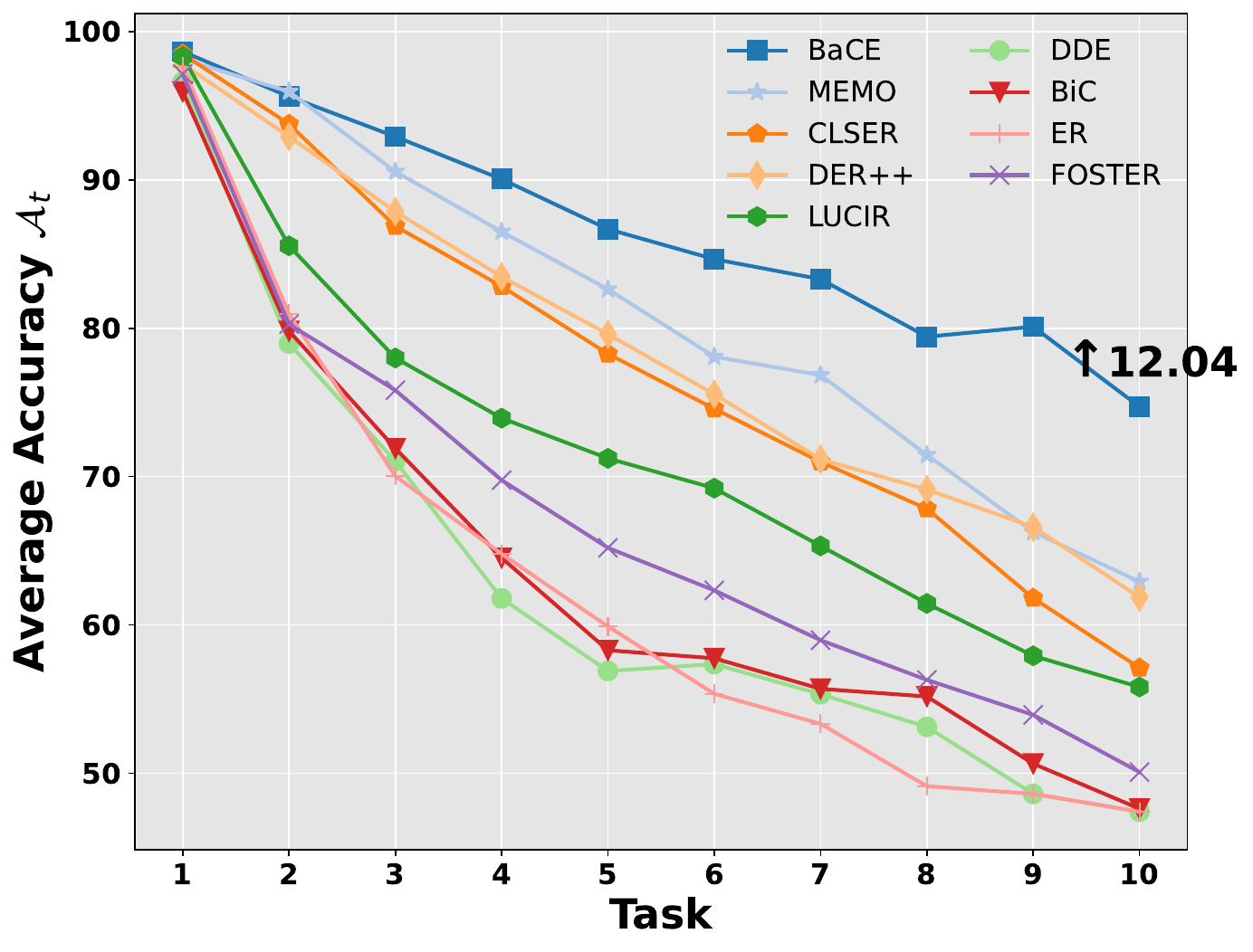}
    }
    \subfloat[10 steps, 500 buffer size]{
        \includegraphics[width=0.49\linewidth]{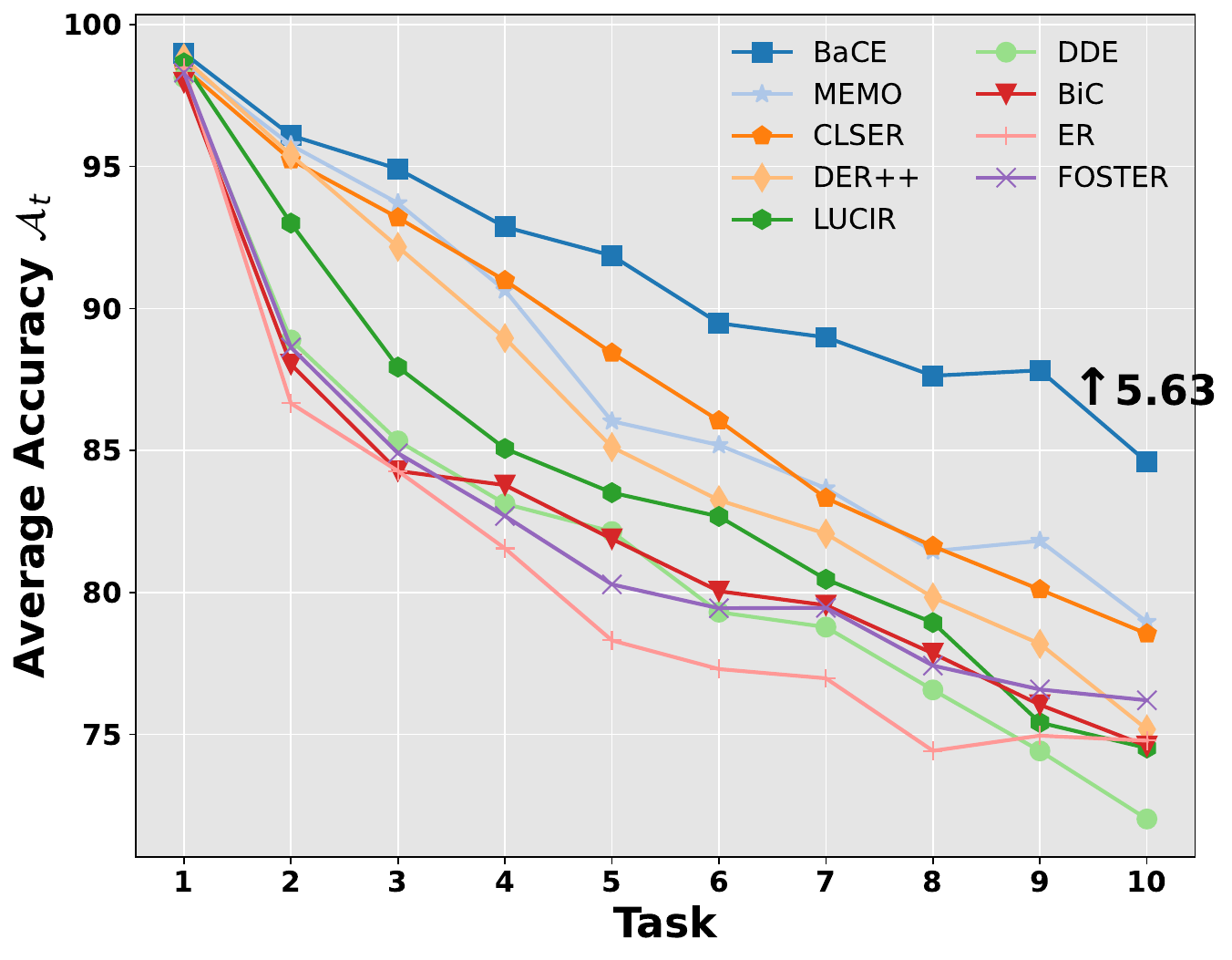}
    }
    \caption{The evolution of average accuracy on split CIFAR-100 with ViT-B/16.}
    \label{fig:evolution_acc_ic}
\end{figure*}

The evolution of average accuracy on split CIFAR-100 with ViT-B/16 is provided in Figure \ref{fig:evolution_acc_ic}.

\subsection{Visualization of the KNN Examples in $\textit{Effect}_{new}$} 

\begin{figure}[htbp]
    \centering
    \includegraphics[width=0.9\linewidth]{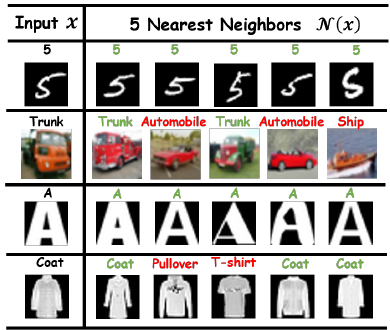}
    \caption{The KNNs in $\textit{Effect}_{new}$ on CIFAR100.}
    \label{fig:case_study_2}
\end{figure}

We provide more KNN examples on CIFAR100 in Figure \ref{fig:case_study_2}.

\subsubsection{Comparison with Existing Methods for Class Imbalance Problem}

\begin{figure*}[htbp]
    \centering
    \subfloat[LUCIR]{
        \includegraphics[width=0.18\linewidth]{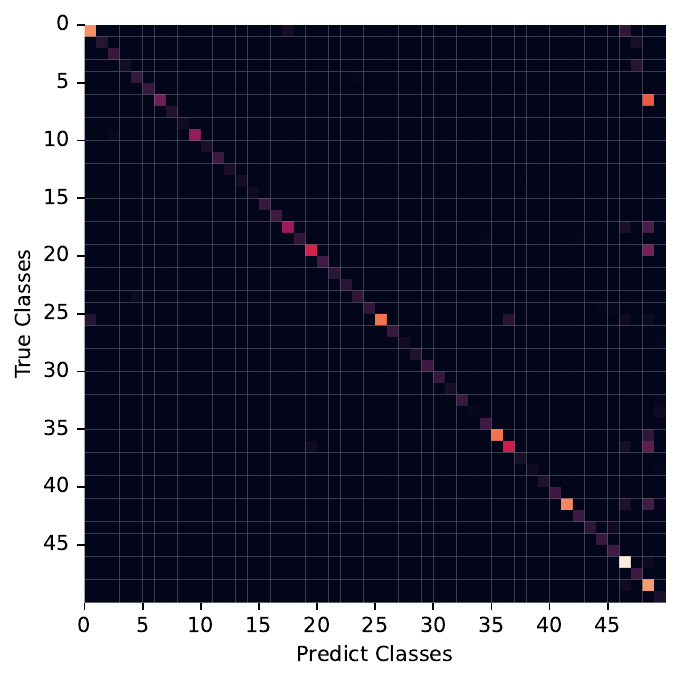}
    }
    \subfloat[IL2M]{
        \includegraphics[width=0.18\linewidth]{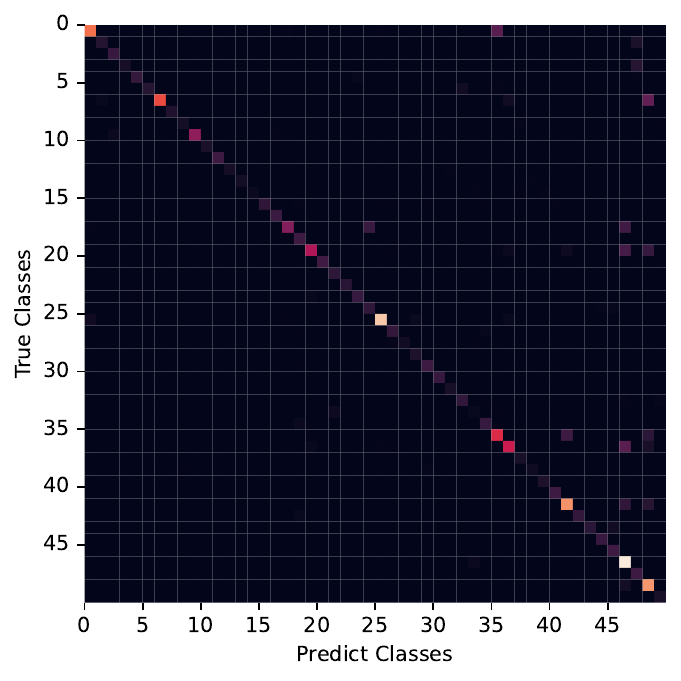}
    }
    \subfloat[BiC]{
        \includegraphics[width=0.18\linewidth]{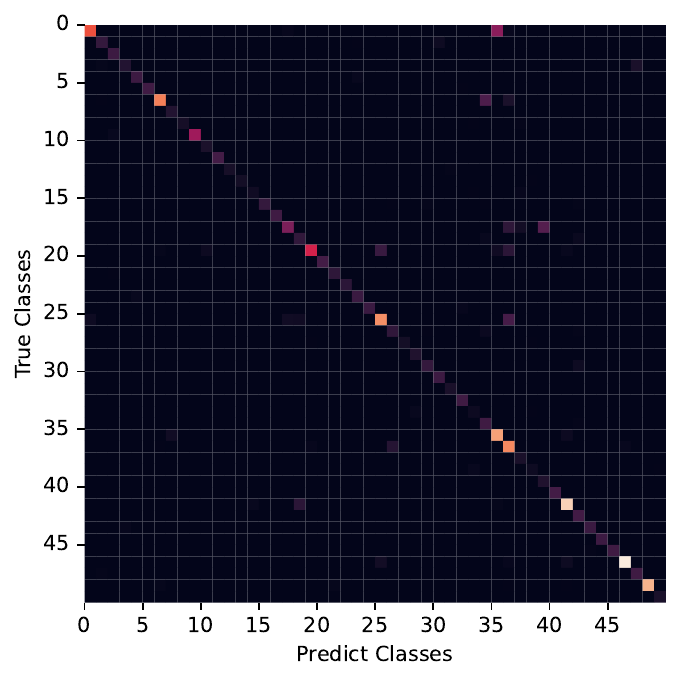}
    }
    \subfloat[BaCE]{
        \includegraphics[width=0.18\linewidth]{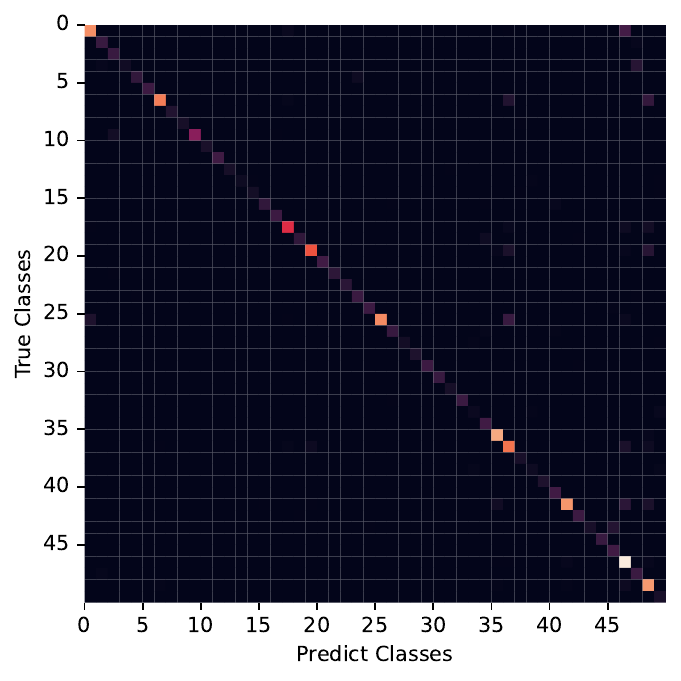}
    }
    \subfloat[BaCE+Bias Correction]{
        \includegraphics[width=0.21\linewidth]{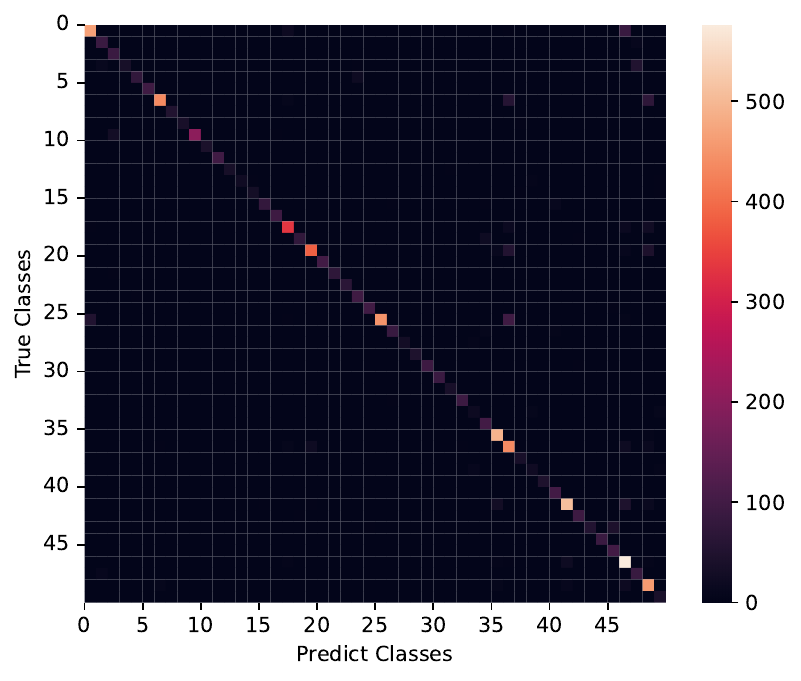}
    }
    \caption{The confusion matrix of different methods on VTAB.}
    \label{fig:confusion_matrix_VTAB}
\end{figure*}

To have a closer look at the class imbalance problem, we provide the confusion matrices of BaCE, BiC, LUCIR, IL2M on VTAB dataset in Figure \ref{fig:confusion_matrix_VTAB}.

\subsection{Extending to Continual Text Classification}

We choose bert-base-cased \cite{devlin-etal-2019-bert,wolf2019huggingface} as the backbone model for all approaches due to its popularity in NLP.
The pretrained weights are downloaded from Huggingface \cite{wolf2019huggingface}.
We train each model for 3 epochs and tune the batch size in \{8, 16\}, the learning rate of the backbone model in \{1e-4, 3e-5, 1e-5\}.
The learning rate of the final linear classifier is set as 1e-3, and the maximum sequence length is set as 128.
In $\textit{Effect}_{new}$, we set the number of neighbours as $K=5$ and the weight $W_0=0.95$.
In $\textit{Effect}_{old}$, we set $\alpha=1$.

We use three publicly available topic classification datasets, including AGNews, DBPedia, and Yahoo \cite{zhang2015character}, which are collected from various domains.
Different from \cite{de2019episodic}, we do not use Yelp and Amazon since their labels are product ratings from 1-5, which simplifies CIL to Task-Incremental Learning.
Besides, we divide DBPedia and Yahoo into two sub-datasets with disjoint label sets, respectively.
In total, we obtained five datasets, including TC AGNews, DBPedia1, DBPedia2, Yahoo1, and Yahoo2.
Since we focus on CIL, we remove the overlapping labels to ensure the label set between datasets is disjoint.
Then, the categories in each dataset are as follows: AGNews (4 classes including: \textit{World}, \textit{Sports}, \textit{Business}, \textit{Sci\_Tech}); DBPedia1 (6 classes including: \textit{Artist}, \textit{Athlete}, \textit{OfficeHolder}, \textit{MeanOfTransportation},\textit{Building}, \textit{NaturalPlace}); DBPedia2 (6 classes including: \textit{Village}, \textit{Animal}, \textit{Plant}, \textit{Album}, \textit{Film}, \textit{WrittenWork}); Yahoo1 (3 classes including: \textit{Society\_Culture},\textit{Health},\textit{Education\_Reference}); Yahoo2 (3 classes including: \textit{Computers\_Internet}, \textit{Family\_Relationships}, \textit{Politics\_Government}).
Following \cite{de2019episodic}, we randomly sample an equal number of training samples for each category.
Concretely, each category contains 28,000 training samples and 2,000 test samples.
We define the task sequence as follows: AGNews$\rightarrow$DBPedia1$\rightarrow$DBPedia2$\rightarrow$Yahoo1$\rightarrow$Yahoo2.

We compare the following strong baselines:
LwF \cite{li2017learning}, EWC \cite{kirkpatrick2017overcoming}, Sparse-ER \cite{de2019episodic}, ER \cite{chaudhry2019tiny}, MBPA \cite{sprechmannmemory}, MBPA++ \cite{de2019episodic}, IDBR \cite{huang2021continual}.

The result of continual text classification is summarized in Table \ref{tab:main_ctc}.
The result indicates that BaCE outperforms not only CIL methods designed for pretrained language models, including MBPA++ and Sparse-ER, but also the task-specific CIL methods designed for text classification, IDBR.

\begin{table}[htbp]
  \centering
  \caption{The results of continual text classification. All methods use pretrained bert-based-cased as the backbone.}
  \resizebox{1.0\linewidth}{!}{
        \begin{tabular}{clccc}
    \toprule
    \textbf{Buffer Size} & \multicolumn{1}{c}{\textbf{Method }} & \textbf{$\mathcal{A}_{last}$ (↑)} & \textbf{FGT (↓)} & \textbf{FWT (↑)} \\
    \midrule
    \multirow{6}[2]{*}{100} & Sparse-ER \cite{de2019episodic} & 63.71  & 34.12  & 51.90  \\
          & ER    & 69.58  & 26.54  & 53.56  \\
          & MBPA \cite{sprechmannmemory} & 61.06  & 29.78  & 29.04  \\
          & MBPA++ \cite{de2019episodic} & 74.11  & 17.64  & 53.28  \\
          & IDBR \cite{huang2021continual} & 80.82  & 10.55  & 52.77 \\
          & \cellcolor{black!10}BaCE & \cellcolor{black!10}\textbf{81.56 } & \cellcolor{black!10}\textbf{9.70 } & \cellcolor{black!10}\textbf{54.04 } \\
    \midrule
    \multirow{6}[2]{*}{500} & Sparse-ER \cite{de2019episodic} & 64.16  & 33.40  & 52.13  \\
          & ER    & 77.44  & 16.62  & 52.60  \\
          & MBPA \cite{sprechmannmemory} & 61.28  & 30.66  & 27.37  \\
          & MBPA++ \cite{de2019episodic} & 78.72  & 12.70  & 53.73  \\
          & IDBR \cite{huang2021continual} & 83.12  & 10.43  & 53.44 \\
          & \cellcolor{black!10}BaCE & \cellcolor{black!10}\textbf{83.53 } & \cellcolor{black!10}\textbf{8.28 } & \cellcolor{black!10}\textbf{54.21 } \\
    \midrule
    $\infty$ & MTL   & 86.59  & /     & / \\
    \bottomrule
    \end{tabular}%
    }
  \label{tab:main_ctc}%
\end{table}%

\subsection{Extending to Continual Named Entity Recognition}

We use bert-base-cased \cite{devlin-etal-2019-bert} as the backbone model because of its popularity.
We train each model for 5 epochs and tune the batch size in \{8, 16\}, the learning rate of the backbone model in \{1e-4, 3e-5, 1e-5\}.
The learning rate of the final linear classifier is set as 1e-3, and the maximum sequence length is set as 256.
In $\textit{Effect}_{new}$, we set the number of neighbours as $K=5$ and the weight $W_0=0.95$.
In $\textit{Effect}_{old}$, we set $\alpha=1$.

We select five commonly-used Named Entity Recognition (NER) datasets, including CoNLL2003 \cite{sang2003introduction}, I2B2 \cite{murphy2010serving}, MIT Restaurant \cite{DBLP:conf/icassp/LiuPCG13}, MIT Movie \cite{DBLP:conf/asru/LiuPWCG13}, OntoNotes5 \cite{hovy2006ontonotes}.
Compared with continual TC, continual NER is a more practical and challenging scenario because the number of classes between tasks and samples between classes is imbalanced.
For example, the number of entities in OntoNotes5 and CoNLL2003 is 14 and 4.
In the training set of OntoNotes5, there are 10922 \textit{DATE} entities but only 282 \textit{LAW} entities.
We define the task sequence as follows: 
OntoNotes5$\rightarrow$MIT Movie$\rightarrow$I2B2$\rightarrow$MIT Restaurant$\rightarrow$CoNLL2003.

For continual NER, we select the following competitive methods:
LwF \cite{li2017learning}, EWC \cite{kirkpatrick2017overcoming}, Sparse-ER \cite{de2019episodic}, ER \cite{chaudhry2019tiny}, MBPA \cite{sprechmannmemory}, MBPA++ \cite{de2019episodic}, ExtendNER \cite{monaikul2021continual}, CFNER \cite{zheng2022distilling}.
For LwF, EWC, Sparse-ER, ER, MBPA, and MBPA++, we use the same setting as in continual text classification.

The result of continual NER is summarized in Table. \ref{tab:main_cner}.
The result indicates that BaCE consistently outperforms generic methods, including LwF, EWC, Sparse-ER, ER, MBPA, and MBPA++, as well as the task-specific methods designed for NER, including ExtendNER and CFNER.

\begin{table}[htbp]
  \centering
  \caption{The results of continual NER. All methods use pretrained bert-based-cased as the backbone.}
  \resizebox{\linewidth}{!}{
    \begin{tabular}{clccc}
        \toprule
    \textbf{Buffer Size} & \multicolumn{1}{c}{\textbf{Method }} & \textbf{$\mathcal{A}_{last}$ (↑)} & \textbf{FGT (↓)} & \textbf{FWT (↑)} \\
   \midrule
    \multirow{7}[2]{*}{100} & Sparse-ER \cite{de2019episodic} & 29.83  & 48.14  & 28.69  \\
          & ER    & 46.49  & 26.70  & 33.26  \\
          & MBPA \cite{sprechmannmemory} & 38.56  & 30.64  & 26.32  \\
          & MBPA++ \cite{de2019episodic} & 45.37  & 23.56  & 32.10  \\
          & ExtendNER \cite{monaikul2021continual} & 51.69 & 22.94 & 37.8 \\
          & CFNER \cite{zheng2022distilling} & 56.16  & 14.61  & 34.87  \\
          & \cellcolor{black!10}BaCE & \cellcolor{black!10}\textbf{58.33 } & \cellcolor{black!10}\textbf{10.97 } & \cellcolor{black!10}\textbf{42.37 } \\
    \midrule
    \multirow{7}[2]{*}{500} & Sparse-ER \cite{de2019episodic} & 30.81  & 46.34  & 46.34  \\
          & ER    & 57.17  & 13.54  & 36.98  \\
          & MBPA \cite{sprechmannmemory} & 38.98  & 28.00  & 26.11  \\
          & MBPA++ \cite{de2019episodic} & 53.96  & 14.50  & 36.96  \\
          & ExtendNER \cite{monaikul2021continual} & 58.06  & 13.69  & 42.60  \\
          & CFNER \cite{zheng2022distilling} & 60.08  & 12.76  & 42.24  \\
          & \cellcolor{black!10}BaCE & \cellcolor{black!10}\textbf{62.18 } & \cellcolor{black!10}\textbf{7.78 } & \cellcolor{black!10}\textbf{43.98 } \\
    \midrule
    $\infty$ & MTL   & 71.62  & /     & / \\
    \bottomrule
    \end{tabular}%
    }
  \label{tab:main_cner}%
\end{table}%


 





\end{document}